**Title Page**

Original Article

# Optimizing Large Language Models for Causality Assessment in Pharmacovigilance: Developing a Performance Metric as Objective for Bayesian Hyperparameter Optimization

Nicole Sonne Heckmann[1*] & Arnault-Quentin Vermillet[1*], Søren Norlin Mølgaard[2], Manuela Del Castillo Suero[1], Lars Melskens[3], Gerard Ompad[4**] & Maurizio Sessa[1**]

[1]Department of Drug Design and Pharmacology, University of Copenhagen, Copenhagen, Denmark.

[2]Safety Operations, Novo Nordisk, Søborg, Denmark.

[3]Clinical Development Centre Denmark, Novo Nordisk, Søborg, Denmark.

[4]Statistical Data Science Lead, Pfizer, Manila, Philippines.

* Equal contribution as first authors.

** Equal contribution as last authors.

**Metrics**

*Manuscript*: 5756

*Abstract*: 242

*References*: 17

*Figures and tables*: 1 Table and 4 Figures

**Corresponding author:**

Maurizio Sessa,

Department of Drug Design and Pharmacology,

University of Copenhagen,

Jagtvej 160, Copenhagen 2100, Denmark

Email: maurizio.sessa@sund.ku.dk

**Abstract**

**Background:** Growing individual case safety report (ICSR) volumes have intensified demand for scalable automated causality assessment. Large Language Models (LLMs) show promise, yet performance on clinically demanding tasks remains suboptimal and inference-time hyperparameter optimization has not been investigated.

**Objective:** To develop a Gaussian Process (GP)-compatible optimization objective and investigate whether temperature optimization improves LLM-expert agreement on Naranjo causality assessment of Food and Drug Administration (FDA) Adverse Event Reporting System (FAERS) ICSRs.

**Methods:** Expert causality assessments were performed on 723 stratified FAERS cases. OpenAI's GPT-5.2 was evaluated using chain-of-thought (CoT) prompting. Four composite metrics were developed: Weighted Cosine Similarity (WCS), Information-Weighted Agreement Score (IWAS), Entropy-Weighted Agreement and Cosine Similarity Score (EWACS), and Consensus-Weighted Cosine Similarity (CWCS) and Bayesian optimization using a GP surrogate with Probability of Improvement (PoI) acquisition was applied across temperature $\in [0, 2]$.

**Results:** GPT-5.2 outperformed prior biomedical LLMs at baseline ($T = 0$), achieving 74.1% agreement on question 5 and 65.4% on question 10 of Naranjo´s algorithm. Entropy analysis identified these as the sole informative optimization targets. Temperature showed no systematic population-level effect ($\beta = 0.002$, $p = 0.959$). EWACS-guided Bayesian optimization improved causality classification agreement from 45.0% to 72.0% (+27 pp), with the largest gain in Doubtful cases (+42.9 pp).

**Conclusion:** EWACS was identified as the optimal GP-compatible metric. The absence of a universal temperature optimum indicates LLM performance is driven primarily by ICSR content, yet case-specific temperature selection produced meaningful improvements, supporting temperature optimization for LLM-assisted pharmacovigilance.

**Keywords:** pharmacovigilance; large language models; causality assessment; Bayesian optimization; Gaussian Process; temperature optimization.

## 1. Introduction

FDA's Adverse Event Reporting System (FAERS) processed more than two million individual case safety reports (ICSRs) in 2025, compared with 781,619 reports in 2011 (1, 2). This growth has intensified the need for more efficient and scalable Pharmacovigilance (PV) methods, particularly for individual-level causality assessment, a labor-intensive process traditionally requiring careful clinical judgment by medical safety experts (3, 4).

Abate et al. (2025) were the first to evaluate general-purpose LLMs for this task, finding that ChatGPT-3.5-turbo achieved 34% adherence to the WHO causality algorithm on 150 COVID-19 vaccine ICSRs from VAERS, with weaknesses in evaluating temporal and biological plausibility and alternative causes (5). Heckmann et al. (2026) followed up using biomedical LLMs with chain-of-thought (CoT) prompting on 150 FAERS ICSRs and achieved 63.3% overall classification agreement with human experts, with question-level agreement on the clinically demanding questions (temporal plausibility, alternative causes, objective evidence) ranging from 52% to 64% (6). Concurrently, Kefala et al. (2026) demonstrated that a general-purpose model could resolve the listedness detection challenge that had consistently limited performance (7).

Despite these improvements, performance on questions requiring genuine clinical judgment remains insufficient for routine PV use (i.e., temporal/biological plausibility, alternative causes, and objective evidence). It has been hypothesized that inference-time hyperparameter optimization may be a viable solution to improve LLM performance on these tasks, but no study has investigated this to date (6). A prerequisite for Bayesian hyperparameter optimization is a GP-compatible performance metric, a single continuous scalar that is sensitive to the aspects of LLM output quality being optimized. Existing measures such as question agreement (binary per question), Likert ratings (ordinal), and cosine similarity (continuous but capturing only reasoning, not score accuracy) each fail in isolation or introduce structural biases specific to the FAERS reporting context (8, 9).

This study addresses this gap by (1) developing four composite GP-compatible optimization metrics, Weighted Cosine Similarity (WCS), Information-Weighted Agreement Score (IWAS), and Entropy-Weighted Agreement and Cosine Similarity Score (EWACS), and Consensus-Weighted Cosine Similarity (CWCS), that jointly capture question agreement and reasoning agreement between LLM and expert assessments; (2) identifying which metrics are best suited to the spontaneous reporting context; and (3) applying Bayesian temperature optimization with a GP surrogate and PoI acquisition to investigate whether temperature optimization improves LLM-expert agreement on Naranjo causality assessment of FAERS ICSRs compared to a baseline configuration.

## 2. Methods

### *2.1 Data Sources and Pre-Processing*

2,577,782 FAERS cases for Q1-Q2 2025 were retrieved from openFDA. Quarterly compressed ASCII releases covering the DRUG, REAC, DEMO, OUTC, THER, and INDI tables were imported into R using *readr* and *data.table* and merged through the unique case identifier (PRIMARYID). Age values were converted to years and records were deduplicated within each table (10). Drug names were normalized and matched to RxNorm concepts (RxCUI) using the *rxnorm* package against the official RXNCONSO.RRF file (UMLS 2025AA). A stratified down-sampling procedure was applied. Each RxCUI was mapped to its Anatomical Therapeutic Chemical (ATC) 4th-level classification and adverse event preferred terms (PTs) were hierarchically mapped to High-Level Group Terms (HLGTs) using MedDRA v27.1. For each unique ATC4-HLGT combination, one representative case was randomly selected (*sample* function, seed = 1234), reducing the dataset to 55,177 FAERS cases covering 4,668 unique drugs and 9,168 distinct PTs across all 27 MedDRA System Organ Classes. For each sampled case, a structured case description was programmatically generated by concatenating all relevant FAERS fields.

### *2.2 Expert Causality Assessment*

Expert causality assessments were performed by four assessors: three MSc Pharmaceutical Sciences candidates (NSH, DL, MK) from the University of Copenhagen and a medical doctor serving as a senior case assessor at Novo Nordisk (SNM). Together they evaluated 723 stratified FAERS cases between September 2025 and June 2026 using the Naranjo Algorithm. For every Naranjo question, individual scores and supporting reasoning texts were recorded in a structured data extraction form. A standard operating procedure (SOP) was developed to ensure consistency (Appendix 1). All disagreements were resolved through plenary consensus. These assessments constitute the human gold standard for model evaluation.

### *2.3 Automated Causality Assessment*

Automated causality assessment was performed using OpenAI's GPT-5.2, accessed via the OpenAI API. All API queries were performed on April 28th, 2026, from Copenhagen, Denmark. A CoT prompting strategy was adopted following Heckmann et al. (2026), who identified CoT as the best-performing technique for this task (6). Each API call comprised a system prompt defining the model's role as a medical safety expert and a user prompt containing the full structured case narrative followed by the respective Naranjo algorithm question (4). The Naranjo Adverse Drug Reaction Probability Scale is one of the most widely used structured algorithms for individual-level causality assessment, and previous studies have identified it as the best-performing algorithm when paired with LLMs for this task (6). For each question, the model was instructed to provide both a score and a reasoning. The exact prompt template is shown in Supplementary Table S1. GPT-5.2 was evaluated at a baseline configuration of T = 0.

### *2.4 Evaluation Metrics and Scoring Framework*

Question agreement was coded as 1 when the LLM and expert numeric scores matched and 0 otherwise. Total Naranjo scores were summed, and the resulting Naranjo causality categories (Definite, Probable, Possible, Doubtful) were compared to evaluate overall consistency.

Reasoning agreement was assessed using cosine similarity between sentence embeddings of the LLM and expert reasoning texts. Both texts were encoded into dense vector representations using the paraphrase-multilingual-mpnet-base-v2 sentence embedding model (11). Cosine similarity was computed between the resulting embedding vectors:

$$\cos(u, v) = \frac{(u \cdot v)}{(||\mathrm{u}|| \cdot ||\mathrm{v}||)} \tag{1}$$

where u and v are the embedding vectors of the two texts, u · v is their dot product, and ||u|| and ||v|| are their respective magnitudes. The score ranges from -1 to 1, where values approaching 1 indicate high semantic alignment. Because contextual sentence embeddings can be anisotropic (occupying a narrow cone of the vector space, which inflates cosine similarity), the embedding space was diagnosed for anisotropy, and an optional Principal Component Analysis (PCA)-whitening correction was implemented. Diagnostics on the present data showed that the chosen contrastively trained model already produced near-isotropic embeddings: cosine similarity between semantically unrelated question reasonings (e.g., question 2 versus question 10) was approximately 0.30-0.40 (an angular separation of roughly 65-70°), indicating that whitening was not required. Whitening was therefore disabled in the primary analysis and retained only as a sensitivity option (Supplementary Figure S1).

### *2.5 Optimization Objective Development*

Four metrics were developed as Bayesian optimization objectives, each reflecting different analytical priorities. All build on the two core components: question agreement and reasoning agreement.

Weighted Cosine Similarity (WCS) combines reasoning similarity and score agreement into a single score per Naranjo question, then averages across all nine:

$$\mathrm{WCS} = \frac{1}{|\,Q\,|} \sum_{q=2}^{10} \left[ \cos(\mathrm{e}_q^{\mathrm{LLM}}, \mathrm{e}_q^{\mathrm{exp}}) - \frac{1 - a_q}{2} \right] \tag{2}$$

where Q is the set of all nine Naranjo questions, $e_q^{LLM}$ and $e_q^{exp}$ are the sentence embedding vectors of the reasoning texts for question q, cos(·) is the cosine similarity, and $a_q$ = +1 on agreement and -1 on disagreement, so the penalty (1 - $a_q$)/2 evaluates to 0 on agreement and 1 on disagreement. The aggregate WCS ranges from -2 to 1, with all nine questions weighted equally.

Information-Weighted Agreement Score (IWAS) uses binary score agreement weighted by normalized Shannon entropy, so that questions with low or zero entropy contribute proportionally less:

$$\text{IWAS} = \sum_{q=2}^{10} [agree(q) \times w(q)], \qquad w(q) = \frac{H(q)}{\sum_{q'} H(q')} \tag{3}$$

where agree(q) is a binary agreement indicator, H(q) is the Shannon entropy of the expert score distribution for question q, and w(q) is the normalized entropy weight. Shannon entropy is defined as:

$$H(q) = -\sum_{s} p_q\,(s) \log_2 p_q(s) \tag{4}$$

where s indexes the possible score values and $p_q(s)$ is the proportion of cases receiving score s on question q. Questions with H(q) = 0 (no score variability) are excluded automatically. IWAS is bounded between 0 and 1.

Entropy-Weighted Agreement and Cosine Similarity Score (EWACS) combines elements of both metrics by retaining the cosine similarity component from WCS but applying entropy-based weighting as in IWAS:

$$\text{EWACS} = \sum_{q=2}^{10} H(q) \times \left[\cos(\mathrm{e}_q^{\mathrm{LLM}},\ \mathrm{e}_q^{\mathrm{exp}}) - \frac{1 - a_q}{2}\right] \tag{5}$$

where H(q) is the raw Shannon entropy in bits (not normalized). Questions with H(q) = 0 are excluded automatically, and each remaining question's per-question WCS contribution is scaled by how informative that question is. The theoretical range of EWACS is [-2.73, 1.37]; the observed range across the dataset was approximately [-0.33, 0.85].

A fourth metric, Consensus-Weighted Cosine Similarity (CWCS), was developed as a corrected, bounded objective designed specifically for Gaussian-Process optimization. Whereas WCS and EWACS subtract a binary agreement penalty from the cosine term, CWCS multiplies a soft (continuous) score-agreement term by a rescaled reasoning-similarity term, so that a high value requires both score and reasoning alignment. The per-question value is $CWCS_{ik} = \lambda_{ik} \times s_{ik}$, where $\lambda_{ik} = 1 - |\hat{y}_{ik} - y_{ik}|$ is the soft score-agreement term obtained from min-max normalized Naranjo scores ($\lambda = 1$ on exact agreement, decreasing with normalized disagreement) and $s_{ik}$ rescales the LLM-expert reasoning cosine similarity for that question from [−1, 1] to [0, 1] via (cosine + 1) / 2. The case-level value is the weighted mean $CWCS_i = \Sigma_k w_k\, CWCS_{ik} / \Sigma_k w_k$, with weights $w_k = M_k / \Sigma_j M_j$ derived from the maximum positive score $M_k$ that each question contributes to the standard Naranjo scale ($\Sigma M_j = 13$), preserving the scale's intrinsic weighting. CWCS is bounded in [0, 1] and continuous in both components, providing a smoother GP response surface than the binary-agreement metrics. Because its cosine term is multiplicatively gated, an optional anisotropy correction of the embedding space (PCA whitening) was implemented as a sensitivity option and disabled in the primary analysis.

$$\mathrm{CWCS} = \sum_{q=2}^{10} w_q \lambda_q \frac{\left(\cos(\mathrm{e}_q^{\mathrm{LLM}}, \mathrm{e}_q^{\mathrm{exp}}) + 1\right)}{2} \quad (6)$$

***2.6 Bayesian Temperature Optimization***

The temperature range explored was T ∈ [0, 2], covering the full range available through the OpenAI API for GPT-5.2. The optimization objective *f(T)* was modelled as a Gaussian Process:

$$f\,(T) \sim \mathcal{GP}(\mu_0,\, k(T, T')) \quad (7)$$

where $\mu_0$ is a constant prior mean function and k(T, T') is the Matérn 5/2 covariance kernel:

$$k(T, T') = \sigma^2 \left(1 + \frac{\sqrt{5}\mid T - T' \mid}{\ell} + \frac{5(T - T')^2}{3\ell^2}\right) \exp\left(\frac{\sqrt{5}\mid T - T' \mid}{\ell}\right) \quad (8)$$

where $\sigma^2$ is the signal variance and l is the lengthscale. GP hyperparameters were fixed: the GP observation noise was specified as a variance, because BoTorch/GPyTorch parameterize the Gaussian likelihood noise as a variance rather than a standard deviation. The noise variance was therefore set to noise_sd² with the intended noise standard deviation (noise_sd) in the range 0.05-0.12 on the bounded [0, 1] metric scale (0.12 for single-case within-case optimization; 0.05-0.08 for across-case experiments), and the lengthscale to 0.10-0.20 throughout. A noise value of 0.35 entered directly as a variance corresponds to a noise standard deviation of approximately 0.59 on a [0, 1] objective, which over-smooths the surrogate; this value was therefore not used. All GP computations used *BoTorch (v0.12.0)* and *GPyTorch (v1.13)*.

The Probability of Improvement (PoI) acquisition function was used to select the next temperature at each iteration:

$$\mathrm{PoI}(T) = \Phi\left(\frac{\mu(T) - f^* - \varepsilon}{\sigma(T)}\right) \quad (9)$$

where $f^*$ is the best performance observed so far, $\mu(T)$ and $\sigma(T)$ are the GP posterior mean and standard deviation at temperature T, $\varepsilon = 0.02$ is the exploration margin (a small value is appropriate because the metric is bounded in [0, 1] and baseline values are already high near 0.7; the previously used value of 0.3 would require an unrealistically large improvement and renders PoI overly conservative), and $\Phi$ is the standard normal cumulative distribution function.

Four optimization experiments were conducted. First, IWAS optimization across a randomly sampled subset of N = 100 cases evaluating questions 2, 5, and 10, preceded by a Phase 1 grid sweep at 10 evenly spaced temperatures across [0, 2] on the same 100 cases (yielding 1,000 observations) to warm-start the GP. Second, a question 5-only IWAS experiment on N = 100 cases to test whether the highest-entropy question showed any temperature dependence in isolation, also warm-started from the Phase 1 grid sweep. Third, EWACS optimization on N = 100 cases using all nine Naranjo questions without a grid warm-start. All three experiments were limited to N = 100 due to the high empirical correlation between IWAS and EWACS (Pearson $r = 0.976$, $p < 0.001$) observed on the pilot dataset, indicating that a full-scale run was unlikely to yield qualitatively different conclusions, while minimizing API costs. The optimization was subsequently repeated using the CWCS objective and the corrected GP parameterization (noise specified as a variance via $noise_sd^2$, with noise_sd 0.05-0.12; $\varepsilon = 0.02$), comprising a CWCS grid sweep (100 cases; 1,000 observations), a within-case CWCS experiment (25 cases evaluated repeatedly under BO-guided temperature selection), and a question 5-only CWCS experiment on all 723 cases. As a methodological robustness check, the acquisition strategy was additionally cross-validated by replacing GP-based Probability of Improvement with a Tree-structured Parzen Estimator (TPE; Optuna) optimizing the same CWCS objective.

### *2.7 Study Outcomes*

The primary outcome was the development of a GP-compatible performance metric for Bayesian optimization of LLM-based causality assessment, defined as the metric producing the most reliable and well-converged optimization behavior. The secondary outcome was the performance of GPT-5.2 under an EWACS-optimized temperature configuration compared to the baseline configuration (T = 0), evaluated against acceptable question agreement thresholds established in consultation with industry collaborators at GSK (Supplementary Table S2).

### *2.8 Statistical Analysis*

Descriptive statistics were used to characterize the FAERS datasets. Continuous variables are reported as means with 95% confidence intervals and medians; categorical variables as counts and percentages. Baseline LLM-expert agreement was assessed by computing question agreement rates and Cohen's kappa per Naranjo question. Shannon entropy was calculated per question with a threshold of 0.3 bits to classify questions as informative or non-informative. Normalized entropy (raw entropy divided by the maximum entropy for the question's number of attainable score categories) was additionally reported, because raw entropy in bits is not directly comparable across questions with different score ranges (Supplementary Table S3). Temperature-metric relationships were assessed using linear regression and LOWESS smoothing. The question 5 disagreement rate across temperature bins was tested for homogeneity using a chi-square test. Metric validity was assessed using independent-samples t-tests comparing mean metric scores between correctly and misclassified cases.

### *2.9 Implementation*

All data acquisition, preprocessing, and statistical analysis was performed in R (v4.5.1) using *readr*, *data.table*, *dplyr*, and *tidyr*. Model deployment, similarity scoring, and Bayesian optimization were implemented in Python 3.11 in Google Colaboratory (Pro subscription; 51 GB RAM; NVIDIA Tesla T4 GPU). Key packages included *PyTorch (v2.6.0)*, *GPyTorch (v1.13)*, *BoTorch (v0.12.0)*, *sentence-transformers (v4.1.0)*, and *openai (v1.77.0) (12, 13)*. A fixed random seed was used throughout (seed = 4230893469047). The study was reported in accordance with the TRIPOD-LLM and CHART reporting guidelines (14-16). The full analysis pipeline (metric implementations, Bayesian and TPE optimization, and figure-generating notebooks) and all supplementary figures and tables referenced below are available in the GitHub project repository and accompanying Supplementary Materials.

## 3. Results

### *3.1 Descriptive Analysis*

Three datasets were used: the full stratified FAERS extract (n = 55,177), the expert-assessed subset (n = 723), and the randomly sampled pilot subset used for the EWACS optimization experiment (n = 100) (Figure 1). Demographic and reporter characteristics were broadly consistent across dataset tiers: female cases outnumbered male cases, the age distribution was right skewed with most cases concentrated in the 50-80-year range, the US was the dominant reporting country, and consumers, medical doctors, and other health professionals consistently represented the largest reporter groups.

Adverse events (AEs) burden was lower in the assessed subset (median 3 PTs per case, mean 4.61 [95% CI: 4.25-4.97]) than in the full dataset (median 5, mean 8.85 [95% CI: 8.74-8.95]). Drug count per case was also substantially lower in the assessed subset: median 1, mean 1.26 (95% CI: 1.14-1.39), with 94.9% of cases involving a single suspect drug, compared to a median of 6 and a mean of 9.36 in the total dataset. All 27 MedDRA System Organ Classes were represented across all three datasets.

### *3.2 Baseline LLM-Expert Agreement*

Score distributions for GPT-5.2 and expert assessors across Naranjo questions 2-10 at baseline (T = 0) are presented in Figure 2. Agreement rates ranged from 63.6% for question 2 to >99.0% for questions 4, 6-9. Questions 4, 6, 7, 8, and 9 met the ≥95% threshold. Questions 7, 8, and 9 showed perfect agreement (100%), with all cases receiving a score of 0 from both assessors, reflecting the consistent absence of toxicological measurements, dose-titration records, and prior reaction history in FAERS reports. Questions 2, 5, and 10 fell below threshold. For question 2, expert assessors scored 0 in 99% of cases due to missing therapy dates, while the LLM frequently scored 2, inferring temporal plausibility from context. For question 5, disagreements were bidirectional across 186 cases. For question 10, the LLM scored 0 in 247 cases where the expert scored 1, reflecting a consistent conservative bias.

GPT-5.2 outperformed MedLlama3-8B (the best-performing model from Heckmann et al.) on all Naranjo questions (Supplementary Figure S2). For clinically demanding questions, agreement improved over Heckmann et al.: question 2 (54.0% to 63.6%), question 3 (80.0% to 87.1%), question 5 (52.7% to 74.1%), and question 10 (63.3% to 65.4%). Overall classification agreement was lower in this study than in Heckmann et al. (45.0% vs. 63.3%), reflecting systematic scoring differences on questions 2 and 10 affecting total score distributions. The mean total Naranjo score was 2.51 for expert assessors and 2.88 for the LLM at baseline (Pearson $r = 0.08$, $p = 0.030$), with the LLM producing a higher proportion of Probable classifications and the expert distribution concentrated in the Possible category.

### *3.3 Metric Characterization*

The full informativeness profile is presented in Table 1. Questions 7-9 showed zero expert entropy across all 723 cases. Question 6 showed an agreement rate of 98.8% and an entropy of 0.000 bits. The two questions with the highest informational content were question 5 ($H = 0.751$ bits, agreement 74.1%, $\kappa = 0.180$) and question 10 ($H = 0.445$ bits, agreement 65.4%, $\kappa = 0.220$). Question 2 showed 63.6% agreement despite low entropy ($H = 0.077$ bits, $\kappa = 0.011$), reflecting a systematic one-directional LLM error rather than case-to-case variability in expert scoring.

Analysis of expert reasoning text uniqueness showed that for questions 2-4 and 6-9, the dominant reasoning template accounted for most texts (uniqueness rates of 0.3%-1.5%), reflecting the consistent absence of therapy dates, dechallenge/rechallenge outcomes, and toxicological data in FAERS. When expert reasoning is near-identical across cases, the cosine similarity component measures proximity to a fixed template rather than case-specific reasoning quality. Question 5 was the exception, with 20.5% unique texts overall, rising to approximately 92% among cases where the expert identified a plausible alternative cause. For questions 2-4 and 6-9, including cosine similarity in an optimization metric therefore adds no independent signal.

IWAS entropy weights concentrated almost entirely on questions 5 (54.9%) and 10 (32.6%), with questions 6-9 contributing 0% and the remaining questions contributing a combined 12.5%. Compared to equal-weighted WCS (mean 0.438), restricting to informative questions only yields a mean of 0.200, reflecting substantial inflation from zero-entropy questions that IWAS resolves. On the 100-case pilot dataset, metric distributions showed that WCS had the narrowest spread (mean 0.452, SD 0.098), IWAS the highest mean (0.771, SD 0.324), and EWACS the widest range, extending into negative territory (mean 0.404, SD 0.439). EWACS and IWAS were highly correlated ($r = 0.976$, $p < 0.001$); WCS showed lower correlations with both (Figure 3). On the same 100-case dataset, CWCS, the bounded multiplicative objective, had a mean of 0.713 (SD 0.085, range 0.500-0.867) and correlated only moderately with the other metrics ($r = 0.66$ with WCS, $r = 0.54$ with EWACS, and $r = 0.44$ with IWAS), consistent with its distinct construction (Supplementary Figure S3 and Supplementary Tables S4). Computed on the full assessed dataset (N = 723), the baseline CWCS was 0.693 (SD 0.092; range 0.428-0.826), with a mean soft score-agreement term of 0.891 and a mean reasoning cosine of 0.563; its population identity, $E[CWCS] = E[\lambda]\cdot E[similarity] + Cov(\lambda, similarity)$, was verified for every question, confirming that the metric rewards the co-occurrence of high reasoning quality and high score agreement (Supplementary Tables S5).

### *3.4 Bayesian Optimization Results*

The Phase 1 grid sweep evaluated IWAS across 100 cases at 10 evenly spaced temperatures spanning [0, 2], yielding 1,000 observations. Mean IWAS was stable across the full range, varying narrowly between 0.648 at $T = 2.0$ and 0.695 at $T = 0.444$ (grand mean = 0.675; Figure 3 - Panel A). Standard deviations were consistent across all grid points (range 0.288-0.306). No systematic trend was apparent at the individual or population level, confirming the absence of an obvious temperature optimum.

IWAS optimization across questions 2, 5, and 10 (N = 100) confirmed no significant association between temperature and IWAS ($\beta = 0.001$, $r = 0.002$, $p = 0.987$). The LOWESS smoother was flat across the full temperature domain, with the identified peak at $T = 1.055$ falling within the confidence interval of the grand mean. Binned means were stable across all temperature bands (grand mean IWAS = 0.719). The maximum achievable IWAS in this experiment was 0.931 rather than 1.0, as only three of the five questions with non-zero entropy were evaluated (Figure 3 - Panel B).

The Q5-only IWAS experiment confirmed that the probability of the LLM disagreeing with the expert on question 5 did not depend on temperature (OLS: $\beta = 0.043$, $p = 0.315$; chi-square homogeneity test across eight bins: $\chi^2 = 9.45$, $df = 7$, $p = 0.222$). The overall Q5 failure rate was 33%. The question 5-only experiment was repeated across all 723 cases using the CWCS objective and reached the same conclusion: temperature was not associated with Q5 agreement (OLS: $\beta = 0.048$, $p = 0.076$; chi-square homogeneity test across eight bins: $\chi^2 = 8.45$, $df = 7$, $p = 0.295$), with a mean Q5-CWCS of 0.326 (Figure 3 - Panel C).

EWACS-guided Bayesian optimization on the 100-case subset also showed no systematic temperature effect ($\beta = 0.006$, $p = 0.895$). The GP posterior was approximately flat with wide credible intervals, and the BO-selected temperature distribution was bimodal, with 45 cases evaluated near T = 0 and 50 near T = 2. This bimodal pattern is the expected signature of an acquisition function initialized without a grid warm-start: under a flat GP prior, PoI explores the boundaries of the temperature range first, so the distribution reflects the search geometry rather than a genuine dual optimum (Figure 4 - Panel A). Despite the absence of a universal optimum, EWACS-guided optimization improved overall causality classification agreement from 45.0% at baseline (T = 0) to 72.0% (+27.0 pp;). The improvement was largest for Doubtful cases (+42.9 pp, from 22.9% to 65.7%) and for Possible cases (+18.5 pp, from 56.9% to 75.4%). The over-representation of Probable classifications at baseline (31 cases vs. 0 from expert assessors) was largely resolved under optimization. Total Naranjo score distributions shifted substantially toward the expert distribution. After EWACS-guided Bayesian optimization, overall classification agreement of 72.0% exceeded the Heckmann et al. benchmark of 63.3% by 8.7 pp (Figure 4 - Panel B). WCS, IWAS and EWACS were significantly higher for correctly classified cases (N = 72) than for misclassified cases (N = 28), WCS: 0.483 vs. 0.373 ($t = 5.78$, $p < 0.001$); IWAS: 0.911 vs. 0.471 ($t = 9.40$, $p < 0.001$); EWACS: 0.588 vs. -0.068 ($t = 9.02$, $p < 0.001$) (Figure 4 - Panel C).

Re-running the optimization with the corrected GP parameterization and the CWCS objective reproduced this flat temperature response (mean CWCS varied across the ten grid temperatures by only 0.023, between 0.541 and 0.564), as did an independent TPE optimizer, confirming that the null population-level temperature effect is robust to the surrogate configuration and to the choice of metric (Supplementary Figure S4; Supplementary Table S5). A within-case experiment further showed a small but significant negative association between temperature and CWCS ($\beta = -0.279$, $r = -0.45$, $p < 0.001$), indicating that for a fixed case, lower temperatures tended to produce marginally higher LLM-expert agreement (Supplementary Figure S5). Repeating the classification comparison with the CWCS objective in place of EWACS yielded a lower overall agreement on the same 100 cases (64.0% with the GP surrogate and 61.0% with the TPE sampler, versus 72.0% for EWACS), consistent with the metric characterization: the EWACS configuration most improved the borderline Doubtful and Possible cases, whereas the CWCS objective, although better behaved as a continuous GP target, aligned less closely with final Naranjo classification because its reasoning-similarity component is noisier for that endpoint. EWACS therefore remained the best-performing metric for classification (Supplementary Figures S6-S8).

## 4. Discussion

This study developed four GP-compatible metrics for Bayesian temperature optimization of LLMs performing causality assessment on FAERS ICSRs using the Naranjo Algorithm and investigated whether temperature optimization improves LLM-expert agreement. At the population level, temperature had no systematic effect on LLM-expert question agreement. Despite this, EWACS-guided Bayesian optimization on a 100-case subset improved causality classification agreement from 45.0% at baseline to 72.0%, a 27 percentage point improvement. These findings together suggest that the relationship between temperature and output quality is case-specific rather than universal: no single temperature performs best for all cases, but directing the search at the individual case level still produces meaningful aggregate improvements.

### *4.1 Why Metric Development Was Necessary*

Before any optimization could run, a GP-compatible metric was needed that was sensitive to the LLM's sampling diversity. Two structural properties of the FAERS dataset complicated this. First, for most Naranjo questions, expert reasonings were near-identical across cases, with uniqueness rates of 0.3%-1.5% for questions 2-4 and 6-9, a direct consequence of the frequent absence of therapy dates, dechallenge and rechallenge outcomes, and toxicological measurements in FAERS. Second, questions 6-9 had zero Shannon entropy across all 723 cases, contributing a combined WCS inflation of 0.238 points relative to the informative-question-only mean.

Neither issue is a flaw in WCS itself. In information-rich settings such as clinical trial safety data, where therapy timelines, dechallenge records, and objective measurements are available, equal weighting and the cosine component both carry meaningful signal. These problems are specific to the spontaneous reporting context. IWAS addresses both issues by replacing equal weighting with normalized entropy weights and using binary question agreement to eliminate template redundancy. However, discarding reasoning agreement entirely is generally not acceptable in regulated environments where explainability is required for AI-assisted processes. EWACS retains cosine similarity but weights it by raw entropy, so each question contributes in proportion to how informative it is. For question 5, the only question with genuinely case-specific expert reasoning, cosine similarity carries real information, making EWACS the most sensitive metric for capturing case-level differences in LLM performance. The limited independent contribution of the cosine component was confirmed directly: expert reasoning texts were near-identical across cases for eight of nine questions (uniqueness 0.3-1.5%), and a diagnostic projection of the embeddings collapsed almost entirely onto a single dimension (first principal component explaining 91.8% of variance), a signature of template-dominated rather than case-specific text. Consistent with this, the multiplicative CWCS objective tracked final Naranjo classification largely through its soft score-agreement term rather than its reasoning-similarity term, indicating that classification performance is driven mainly by score agreement. This suggests that, where the downstream target is classification, an additive formulation that makes the score-agreement and reasoning-similarity contributions explicit (for example, $\sigma\lambda + (1 - \sigma) \times$ cosine, with σ tuned by sensitivity analysis) may be preferable to a multiplicative one and is a natural refinement of the metric framework.

### *4.2 Null Temperature Effect and What It Means*

The consistent finding across all four metrics and both datasets that temperature has no systematic population-level effect is itself a substantive result. It indicates that whether the model correctly identifies a plausible alternative cause (question 5) or confirms objective evidence (question 10) is driven primarily by what is documented in the case narrative, not by the stochasticity of output generation. A model that identifies a clear alternative cause will score -1 regardless of whether it runs at T = 0.1 or T = 1.5.

The 27 pp improvement under EWACS-guided optimization therefore does not come from finding a universally better temperature, but from identifying case-specific temperatures that lead the model toward reasoning that better matches the expert's judgment. This explains why the improvement was most pronounced for Doubtful cases (+42.9 pp), where deterministic generation at T = 0 produced systematic biases: the LLM frequently assumed temporal plausibility on question 2 and was reluctant to assume objective evidence had been obtained on question 10, pushing total scores upward and causing many Doubtful cases to be misclassified as Possible or Probable. Temperature variation allowed the model to explore alternative reasoning paths for these borderline cases. It is also worth noting that the EWACS optimization ran without a Phase 1 grid warm-start, suggesting that the 27 pp improvement may be a conservative estimate.

### *4.3 Comparison with Prior Work*

Abate et al. (2025) evaluated ChatGPT-3.5-turbo and Gemini 1.5-pro on 150 VAERS cases using the WHO AEFI causality algorithm, finding an overall adherence of 34% with particularly poor performance on temporal plausibility, alternative causes, and objective evidence (5). Direct numerical comparison with the present study is inappropriate, as these studies used different causality algorithms (WHO vs. Naranjo) and different reporting systems (VAERS vs. FAERS).

Heckmann et al. (2026) applied the same Naranjo algorithm and CoT prompting strategy to 150 FAERS ICSRs, with the best-performing biomedical model (MedLlama3-8B with CoT) achieving 63.3% overall classification agreement and 52.7%-63.3% question-level agreement on questions 2, 5, and 10 (6). GPT-5.2, evaluated on the same algorithm and prompting approach, outperformed this benchmark at baseline across all Naranjo questions, and after EWACS-guided optimization achieved 72.0% classification agreement, exceeding Heckmann et al. by 8.7 pp. This pattern is consistent with the broader finding from Kefala et al. (2026), where a general-purpose model resolved the listedness challenge that biomedical LLMs had failed to address (7). Taken together, these findings suggest that the performance gap that domain-specific training was designed to close has been largely overtaken by broader advances in general-purpose LLM capability.

### *4.4 The Question of What Counts as Good Agreement*

Arimone et al. (2007) found very poor overall inter-expert agreement among five senior PV experts independently assessing 31 drug-event combinations ($\kappa = 0.05$), with the weakest agreement on search for non-drug-related causes ($\kappa = 0.12$) and risk factors for drug causation ($\kappa = 0.09$), criteria that map directly onto question 5 of the Naranjo Algorithm (17). The question-level agreement observed for question 5 (74.1%) and question 10 (65.4%) may therefore reflect the genuine difficulty of these tasks rather than model limitations, and both values already exceed the >60% acceptable thresholds established in consultation with GSK. The 72.0% post-optimization classification agreement should be interpreted against this benchmark: if senior PV experts themselves struggle to agree on the same cases, expecting an LLM to match a single expert panel perfectly is not a realistic target. A more practical question is whether the model performs well enough to function as a first-pass tool, where clear-cut cases are handled automatically and ambiguous ones are referred to a human assessor.

### *4.5 Future Directions*

A relevant next step would be to investigate what case-level characteristics drive the case-specific temperature optima identified by Bayesian optimization, and which ICSR features are associated with low agreement. Understanding these patterns would allow identification of which cases are suitable for automated assessment and which should be referred to human experts, enabling a hybrid decision-support workflow for use by medical safety experts and MAHs. Several extensions of the framework are also worth investigating; extending optimization to other inference-time hyperparameters (top-p, or multi-parameter search); replication on other LLMs to test generalizability; and applying the framework to clinical trial safety data, where therapy timelines, dechallenge records, and objective measurements are available, allowing WCS to be evaluated across all nine Naranjo questions. A further direction is the use of AI agents or multi-agent systems, where a single LLM equipped with retrieval tools could actively query product labels to fill FAERS information gaps.

### *4.6 Limitations*

The most fundamental limitation is the information content of FAERS reports. Therapy dates, dechallenge and rechallenge outcomes, drug concentration measurements, and prior reaction histories are consistently absent, effectively reducing the optimization target to questions 5 and 10. This is an inherent feature of spontaneous reporting data rather than a methodological shortcoming, but it meant that the full potential of WCS could not be tested. All experiments were conducted using GPT-5.2; temperature sensitivities identified here are specific to this model and optimal configurations should not be assumed to transfer directly to other LLMs, though the methodological framework can be applied to any model with a configurable temperature parameter. Finally, this study optimized temperature alone, holding top-p, top-k, frequency penalty, and presence penalty at their default values.

## 5. Conclusion

This study developed four GP-compatible metrics, WCS, IWAS, EWACS and CWCS, for Bayesian temperature optimization of LLMs performing causality assessment on FAERS ICSRs using the Naranjo Algorithm. Entropy analysis showed that only questions 5 and 10 carried meaningful signals in the spontaneous reporting context, motivating entropy-weighted metric designs that concentrate the optimization objective on the questions that are informative. EWACS was identified as the optimal metric for this setting, retaining reasoning assessment where it added genuine information while down-weighting questions where expert responses were near-identical across cases.

Temperature had no systematic population-level effect on LLM-expert agreement, indicating that whether the model correctly identifies an alternative cause or confirms objective evidence depends primarily on what is documented in the case narrative. Despite this, case-specific temperature optimization using EWACS improved causality classification agreement from 45% to 72% (+27 pp), demonstrating that directing the search at the individual case level can correct systematic reasoning biases even in the absence of a universal optimum. GPT-5.2 also outperformed the best biomedical LLM from prior work at baseline across all Naranjo questions, suggesting that the latest general-purpose models have largely closed the gap that domain-specific training was designed to address. The framework is model-agnostic and can be extended to other hyperparameters and richer data sources.

## 6. Ethics, Funding, and Conflict of Interest

This study used only publicly available, de-identified data from FAERS and did not involve human participants, tissue, or identifiable personal data. Ethical approval was therefore not required. No funding was received for this study. The authors declare no conflicts of interest.

## 7. Authors' Contribution

NSH and MS designed the study, developed the optimization metrics, performed all data acquisition, preprocessing, model deployment, and statistical analysis, and wrote the manuscript. MDC assisted with data processing and writing of the manuscript. AQV and GO provided academic supervision and critical review of the manuscript. SNM, and LM provided industry supervision, contributed to study design, and reviewed the manuscript. MS conceived the study, provided senior academic supervision, and critically revised the manuscript for important intellectual content. All authors approved the final version.

**8. Acknowledgement**

The authors gratefully acknowledge Gregory Powell and Andrew Bate (GSK) for their valuable insights and constructive discussions during the development of this work.

**Table 1.** Informativeness Profile of Naranjo Questions 2-10 Based on Expert Assessments

| Naranjo Question | Expert Entropy (bits) | Question Agreement (%) | Cohen's κ | Informative (y/n) |
|---|---|---|---|---|
| 2 | 0.077 | 63.6% | 0.011 | n |
| 3 | 0.079 | 87.1% | 0.098 | n |
| 4 | 0.015 | 98.8% | -0.002 | n |
| 5 | 0.751 | 74.1% | 0.180 | y |
| 6 | 0.000 | 98.8% | 0.000 | n |
| 7 | 0.000 | 100.0% | N/A | n |
| 8 | 0.000 | 100.0% | N/A | n |
| 9 | 0.000 | 100.0% | N/A | n |
| 10 | 0.445 | 65.4% | 0.220 | y |

***Legend:*** *Questions are classified as informative (y) if entropy exceeded 0.3 bits. Cohen's κ is N/A where all cases received the same expert score.*

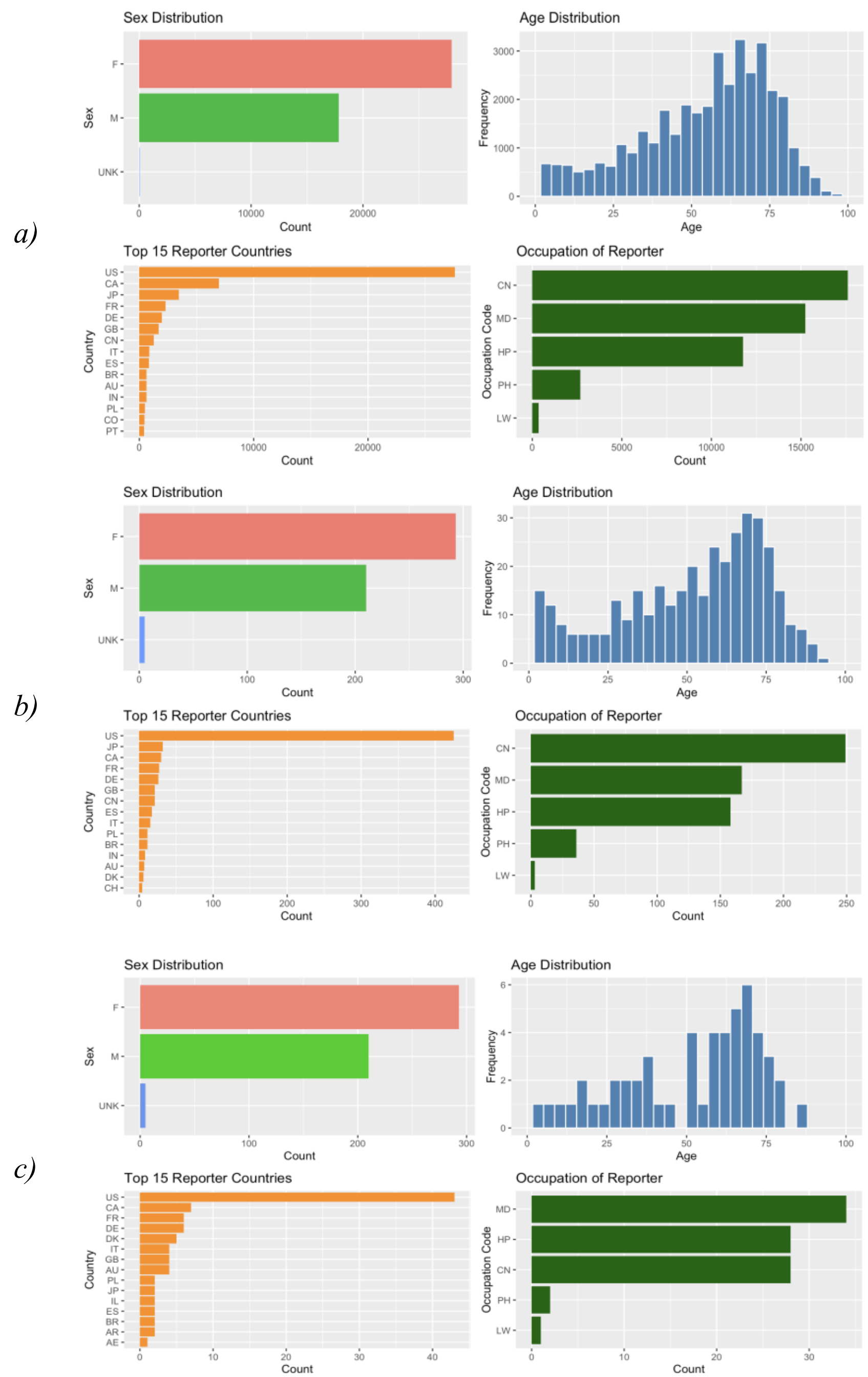

**Figure 1**. Demographic and Reporter Characteristics of FAERS Cases. ***a) the full stratified FAERS extract (n = 55,177); b) the expert-assessed subset (n = 723); c) the randomly sampled pilot subset used for the EWACS optimization experiment (n = 100). Legend****: Bar charts illustrating the distribution of sex, age, reporter country, and reporter occupation across the full stratified dataset (n = 55,177), the expert-assessed subset (n = 723), and the randomly sampled subset (n = 100). Age is reported in years. Countries are represented by ISO 3166-1 alpha-2 codes. Occupation codes: CN = consumer; MD = medical doctor; HP = other health professional; PH = pharmacist; LW = lawyer. UNK = unknown. Only the top 15 reporting countries are displayed.*

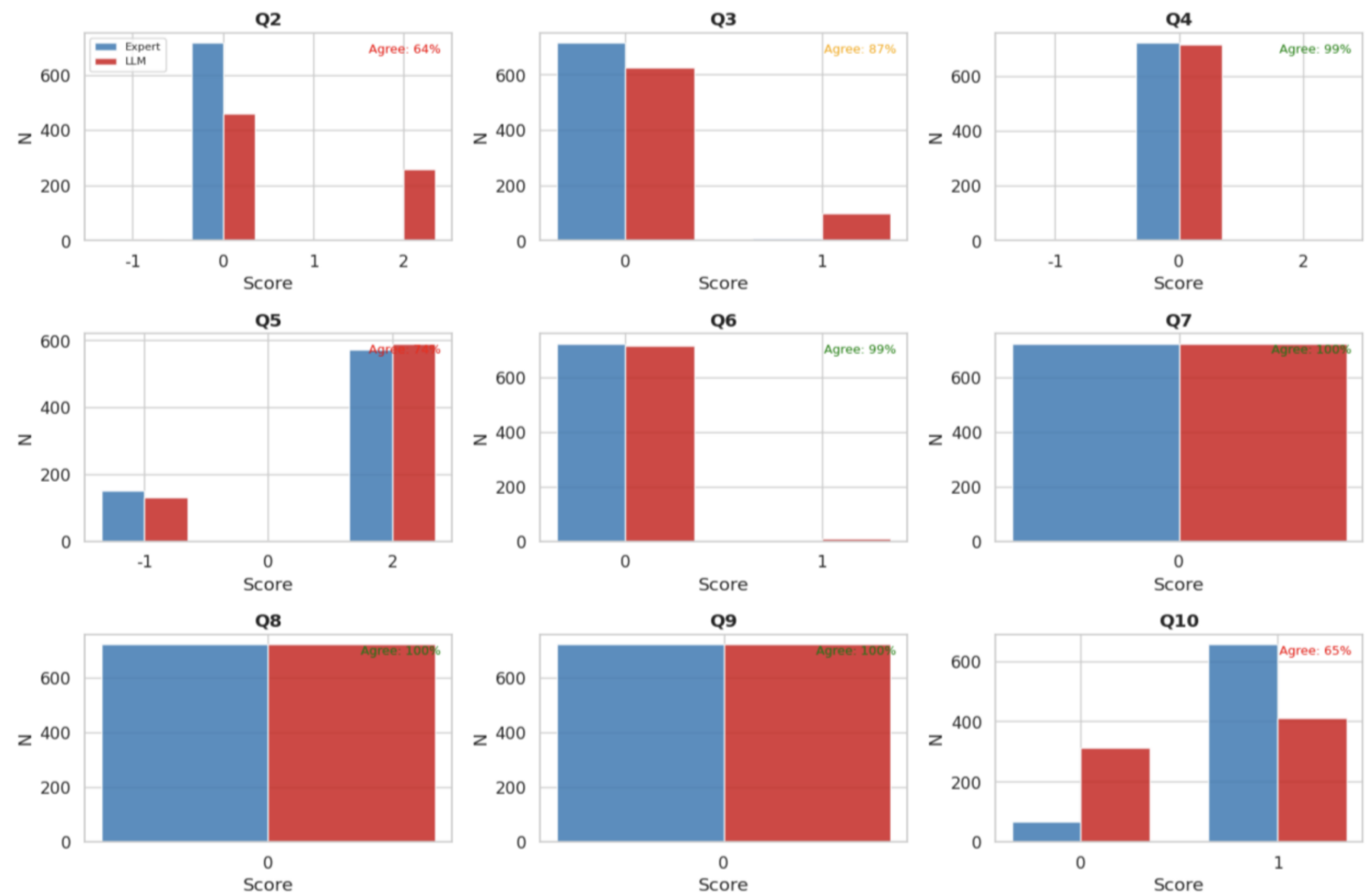

**Figure 2**. Score Distributions per Naranjo Question: Expert vs. Baseline LLM. *Legend: Each panel shows the number of cases receiving each possible score for that question. The agreement rate between expert and LLM is displayed in the top right corner of each panel, colored green (≥ 95%), orange (75-94%), or red (< 75%) to indicate the level of agreement.*

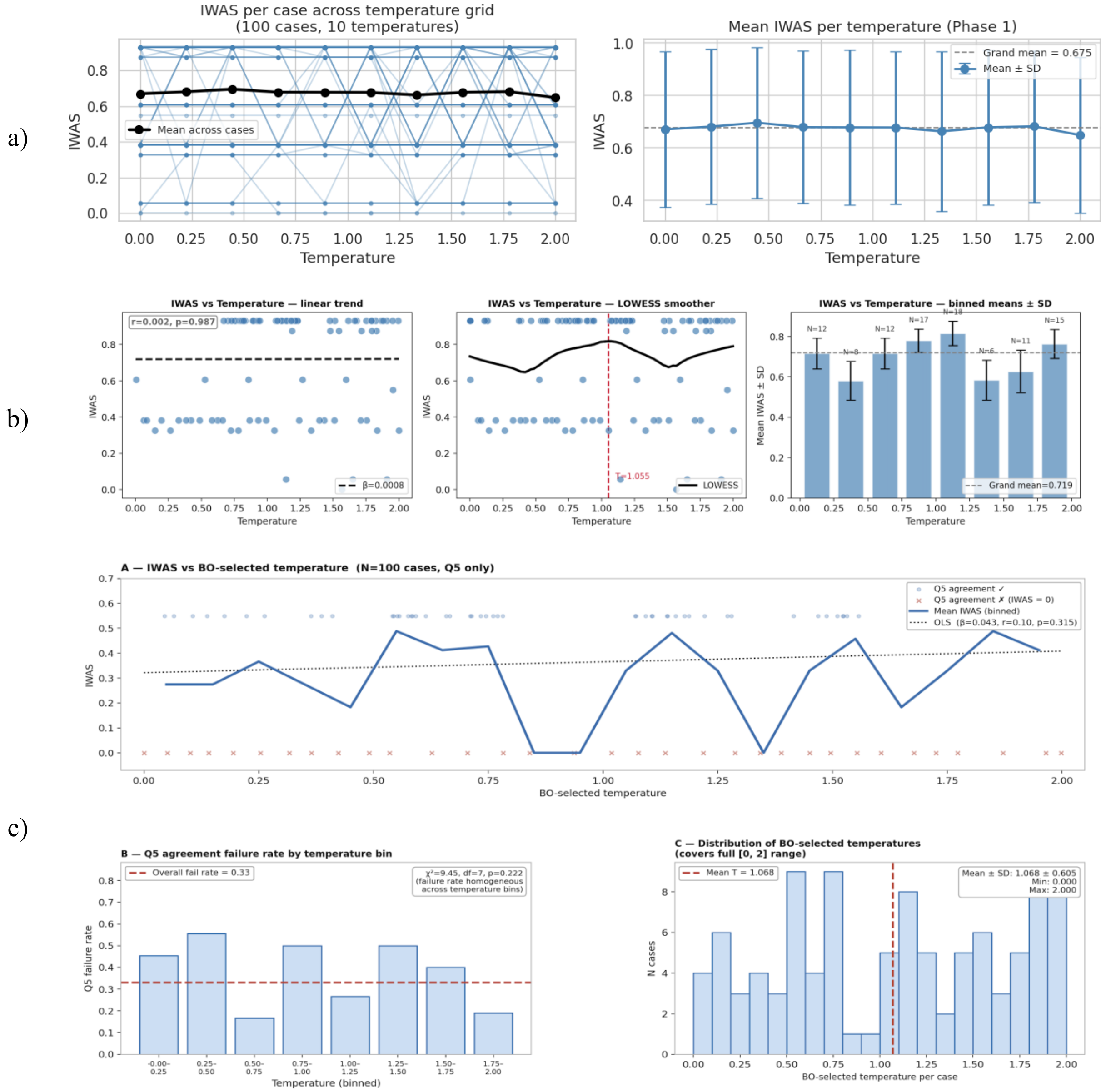

**Figure 3.** Metric comparisons, temperature sensitivity, and LLM causality classification. ***(a) Temperature Grid Sweep:*** *The left panel illustrates individual case IWAS trajectories across 10 evenly spaced grid temperatures (light blue lines), with the overall mean shown in black. The right panel displays the mean IWAS ± SD per grid temperature point; the dashed horizontal line represents the grand mean IWAS across all temperatures and cases (0.675). A summary table below the panels reports the mean IWAS, standard deviation, and number of cases at each evaluated temperature.* ***(b) IWAS vs. LLM Sampling Temperature:*** *The left panel features a scatter plot with a linear trend line and regression statistics. The middle panel displays a scatter plot with a LOWESS smoother, where the dashed red line marks the peak. The right panel presents the mean IWAS ± SD across eight temperature bins, with the grand mean shown as a dashed horizontal line.* ***(c) Bayesian Optimization Temperature Sensitivity using IWAS (Q5 Only):*** *Panel A shows individual IWAS values (circles = Q5 agreement, crosses = Q5 disagreement) and binned means across BO-selected temperatures, with an OLS trend line. Panel B illustrates the Q5 agreement failure rate across eight temperature bins; the dashed red line marks the overall failure rate of 0.33, while a chi-square test confirms homogeneity across bins. Panel C displays the distribution of BO-selected temperatures across the 100 cases.* ***Abbreviations****: BO, Bayesian Optimization; IWAS, Information-Weighted Agreement Score; LLM, Large Language Model; SD, Standard Deviation; WCS, Weighted Cosine Similarity.*

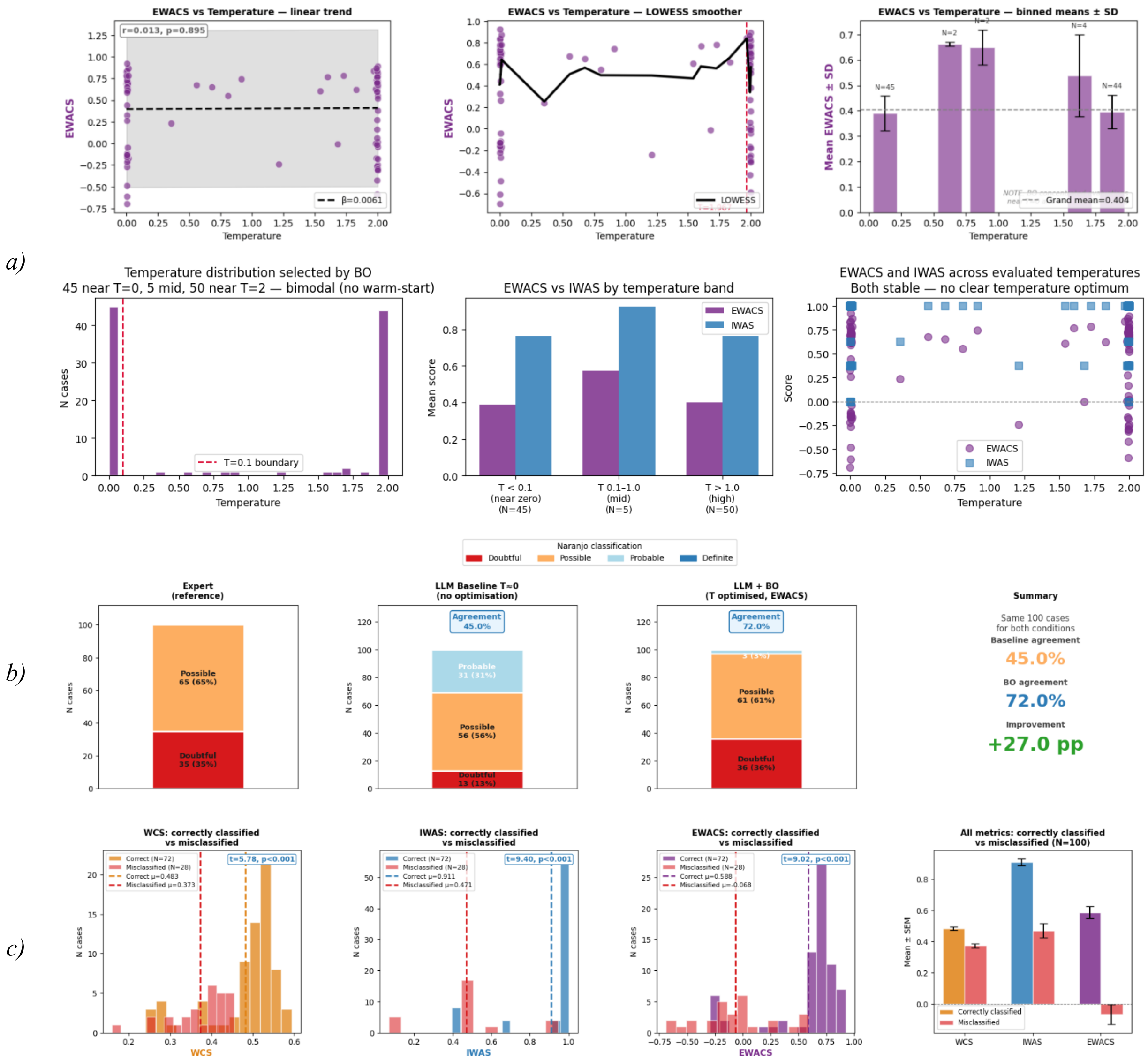

**Figure 4.** Evaluation of Naranjo score distributions, sampling temperature sensitivity, and causality classification metrics. ***(a) EWACS vs temperature:*** *Top row: OLS regression (β = 0.006, r = 0.013, p = 0.895), LOWESS smoother, and binned means ± SD (grand mean = 0.404) confirming no systematic temperature-EWACS relationship. Bottom row: Distribution of BO-selected temperatures (bimodal, T ≈ 0 and T ≈ 2); mean EWACS and IWAS by temperature band (T < 0.1, N = 45; T 0.1-1.0, N = 5; T > 1.0, N = 50); individual EWACS and IWAS scores by temperature.* ***(b) Naranjo Classification: Expert vs. LLM Baseline vs. Bayesian Optimized LLM:*** *Stacked bar charts compare Naranjo causality classification distributions for the expert reference, baseline LLM (T = 0), and EWACS-optimized LLM across the same 100 cases. Overall classification agreement is displayed above each LLM bar. Categories are defined as Doubtful (≤ 0), Possible (1-4), Probable (5-8), and Definite (> 8)* ***(b) WCS vs. IWAS vs. EWACS: Three-Way Comparison:*** *Top row shows pairwise scatter plots of WCS, IWAS, and EWACS across 100 cases, with Pearson correlations and linear trend lines. Bottom left displays overlaid score distributions, where the dashed line marks a score of 0. Bottom middle details the mean per-question contribution to each metric across all nine Naranjo questions. Bottom right presents the mean ± SD for each metric across the 100 cases.* ***(c) Metric Scores for Correctly Classified vs. Misclassified Cases Across All Three Metrics:*** *The left panels show the distribution of WCS, IWAS, and EWACS scores for correctly classified (N = 72) and misclassified (N = 28) cases, including group means (dashed lines) and t-test statistics. The rightmost panel summarizes the mean ± SEM per metric across both groups.*

**SUPPLEMENTARY MATERIAL**

**Supplementary Table S1.** Chain-of-Thought (CoT) Prompt Template Operationalizing the Naranjo Algorithm

| | **Prompt** |
|---|---|
| **General Instructions** | *System prompt:* You are a pharmacovigilance expert. Assess the following case for the drug '[drug name]' and event '[adverse event]'.<br><br>*User prompt:* CASE NARRATIVE: [structured case narrative] TASK: [Naranjo question prompt] Return your answer as: 'Score: <X>' |
| **Naranjo Questions** | Q2: Did the adverse event appear after the suspected drug was administered?<br>Score: 2 for Yes, -1 for No, 0 for Do not know.<br><br>Reasoning: Given the reported timeline and the known route of administration, consider whether the adverse event appeared after the drug was administered. Analyze if the timing is consistent with expected adverse reactions. |
| | Q3: Did the adverse event improve when the suspected drug was discontinued, or a specific antagonist was administered?<br>Score: 1 for Yes, 0 for No/Do not know.<br><br>Reasoning: Search for dechallenge data. If absent, explain how this affects causality evidence. |
| | Q4: Did the adverse event reappear when the suspected drug was readministered?<br>Score: 2 for Yes, -1 for No, 0 for Do not know.<br><br>Reasoning: Search for rechallenge data. If not available, explain how this affects the analysis. |
| | Q5: Are there alternative causes, based only on the information explicitly provided in the case, that could on their own cause the adverse event?<br>Score: -1 for Yes, 2 for No, 0.[*]<br><br>Reasoning: Consider concomitant medications, underlying diseases, infections, procedures. If none identified, note this supports causality. |
| | Q6: Did the adverse event reappear when a placebo was given?<br>Score: -1 for Yes, 1 for No, 0 for Do not know.<br><br>Reasoning: Search for placebo data. If absent, reflect on the impact on causality. |
| | Q7: Was the suspected drug detected in any body fluid in toxic concentrations?<br>Score: 1 for Yes, 0 for No/Do not know.<br><br>Reasoning: Search for toxicology data. If absent, discuss impact on causality score. |
| | Q8: Was the adverse event more severe when the dose was increased or less severe when decreased? |

| | Prompt |
|---|---|
| | Score: 1 for Yes, 0 for No/Do not know.<br><br>Reasoning: Search for dose-change data. Analyze whether symptoms varied with dosing. |
| | Q9: Did the patient have a similar adverse event to the same or similar drugs previously?<br>Score: 1 for Yes, 0 for No/Do not know.<br><br>Reasoning: Consider prior reactions to the same drug or pharmacological class. |
| | Q10: Was the adverse event confirmed by any objective evidence?<br>Score: 1 for Yes, 0 for No/Do not know.<br><br>Reasoning: Determine whether the event is a medical sign (requires objective validation) or a symptom (patient-reported). If a medical sign, assume appropriate diagnostic evaluation was performed even if not explicitly stated, and score Yes. |

[*] *The option of "Do No Know" was omitted for question 5 to align with expert scoring. CoT = Chain-of-thought.*

**Supplementary Table S2.** Acceptable Question Agreement and Rationale for each Naranjo Question.

| Naranjo Question | Acceptable Question Agreement | Rationale |
|---|---|---|
| 2 | >95% | If the event occurred before exposure, it would not be an ICSR; Can be derived from structured ICSR fields |
| 3 | >95% | Can be derived from structured ICSR fields |
| 4 | >95% | Can be derived from structured ICSR fields |
| 5 | >60% | Involves other drugs, medical history, or patient factors |
| 6 | N/A | Placebo use is irrelevant in post-marketing reports |
| 7 | N/A | Concentrations rarely reported |
| 8 | >95% | Can be derived from structured ICSR fields |
| 9 | >60% | Prior exposures seldom documented; may appear in free text |
| 10 | >60% | Confirmation may only be in free text |

***Legend**: Acceptable agreement thresholds and supporting rationale for each of the ten Naranjo Algorithm questions, established in consultation with industry collaborators at GSK. ICSR = Individual Case Safety Report.*

**Supplementary Table S3.** Informativeness Profile of Naranjo Questions 2-10, Including Normalized Entropy.

| Question | Expert entropy (bits) | Normalized entropy | Agreement (%) | Cohen's κ | Informative |
|---|---|---|---|---|---|
| Q2 | 0.077 | 0.049 | 63.6 | 0.011 | No |
| Q3 | 0.079 | 0.079 | 87.1 | 0.098 | No |
| Q4 | 0.015 | 0.015 | 98.8 | −0.002 | No |
| Q5 | 0.751 | 0.474 | 74.1 | 0.180 | Yes |
| Q6 | 0.000 | 0.000 | 98.8 | 0.000 | No |
| Q7 | 0.000 | 0.000 | 100.0 | N/A | No |
| Q8 | 0.000 | 0.000 | 100.0 | N/A | No |
| Q9 | 0.000 | 0.000 | 100.0 | N/A | No |
| Q10 | 0.445 | 0.445 | 65.4 | 0.220 | Yes |

***Legend:*** *Normalized entropy is the raw Shannon entropy divided by the maximum entropy attainable for that question's number of score categories, allowing comparison across questions with different score ranges. Questions are classified as informative when raw entropy exceeds 0.3 bits. κ is not defined where all cases received the same expert score.*

**Supplementary Table S4.** Four-Way Comparison of the Optimization Metrics on the 100-Case Subset: Distribution Summaries (top) and Pairwise Pearson Correlations (bottom).

| Metric | Mean | SD | Min | Max |
|---|---|---|---|---|
| WCS | 0.452 | 0.098 | 0.159 | 0.594 |
| IWAS | 0.771 | 0.324 | 0.000 | 1.000 |
| EWACS | 0.404 | 0.439 | −0.694 | 0.923 |
| CWCS | 0.713 | 0.085 | 0.500 | 0.867 |

*Pearson correlations (r):*

| | WCS | IWAS | EWACS | CWCS |
|---|---|---|---|---|
| WCS | 1.000 | 0.761 | 0.839 | 0.664 |
| IWAS | 0.761 | 1.000 | 0.976 | 0.442 |
| EWACS | 0.839 | 0.976 | 1.000 | 0.543 |
| CWCS | 0.664 | 0.442 | 0.543 | 1.000 |

***Legend:*** *CWCS is bounded and narrowly distributed; it correlates only moderately with the other metrics, reflecting its distinct multiplicative construction.*

**Supplementary Table S5.** Phase 1 CWCS Temperature Grid Sweep (100 cases; Questions 2, 5, 10; 1,000 Observations).

| Temperature | Mean CWCS | SD | SE | N |
|---|---|---|---|---|
| 0.000 | 0.5448 | 0.1727 | 0.0173 | 100 |
| 0.222 | 0.5541 | 0.1635 | 0.0164 | 100 |
| 0.444 | 0.5639 | 0.1676 | 0.0168 | 100 |
| 0.667 | 0.5634 | 0.1739 | 0.0174 | 100 |
| 0.889 | 0.5450 | 0.1750 | 0.0175 | 100 |
| 1.111 | 0.5532 | 0.1726 | 0.0173 | 100 |
| 1.333 | 0.5506 | 0.1827 | 0.0183 | 100 |
| 1.556 | 0.5411 | 0.1615 | 0.0161 | 100 |
| 1.778 | 0.5433 | 0.1627 | 0.0163 | 100 |
| 2.000 | 0.5551 | 0.1664 | 0.0166 | 100 |

***Legend**: Mean CWCS varies by only 0.023 across the full temperature range (best at $T = 0.444$), confirming the absence of a systematic population-level temperature effect under the corrected GP parameterization.*

**Supplementary Table S6.** Naranjo Causality Classification Agreement on the Same 100 cases, Comparing Baseline (T = 0) with Case-Specific Temperature Optimization under Three Configurations.

| Condition | Overall agreement | Doubtful agreement | Possible agreement |
|---|---|---|---|
| Baseline (T = 0) | 45.0 | 22.9 | 56.9 |
| EWACS-optimized (GP) | 72.0 | 65.7 | 75.4 |
| CWCS-optimized (GP) | 64.0 | 77.1 | 56.9 |
| CWCS-optimized (TPE) | 61.0 | 80.0 | 50.8 |

***Legend***: EWACS-guided optimization yields the highest overall agreement; the CWCS objective, although a better-behaved continuous GP target, aligns less closely with the final classification.

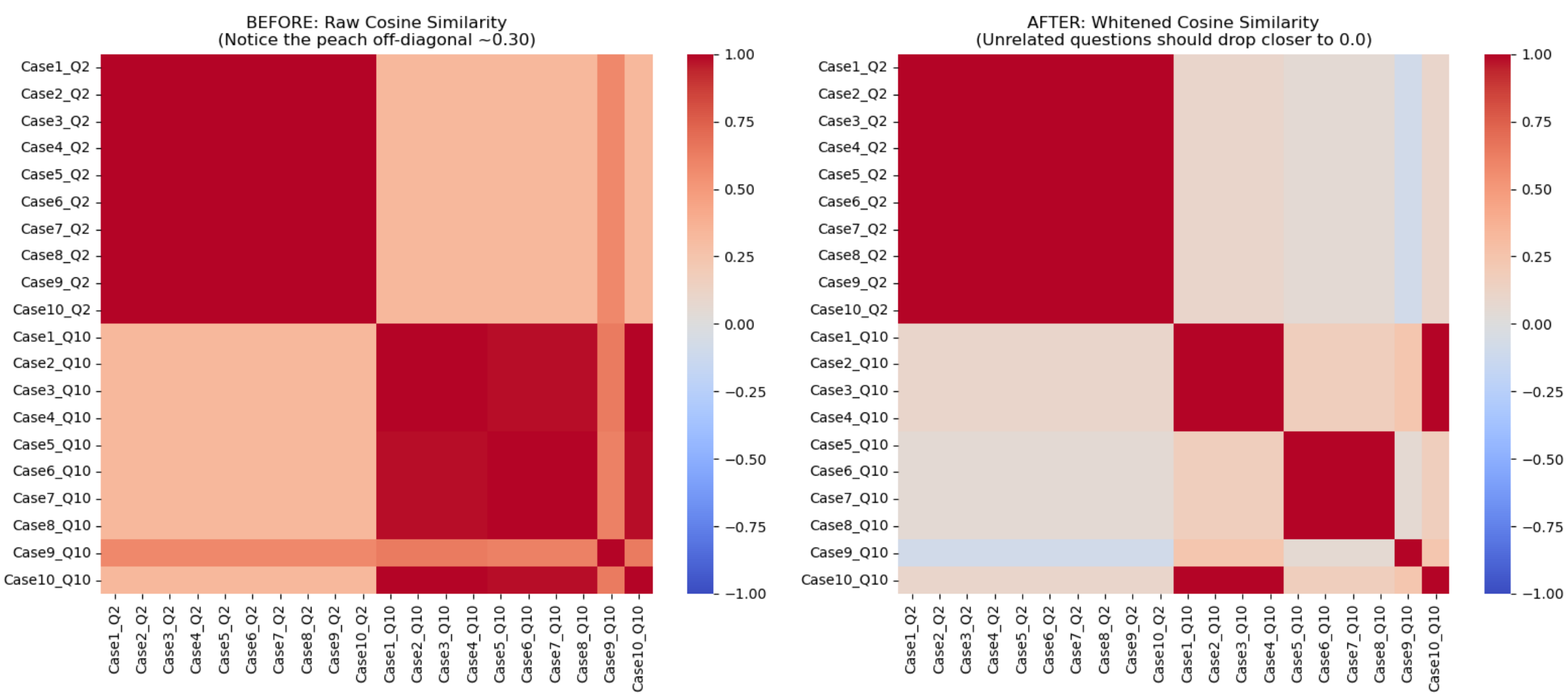

**Supplementary Figure S1.** Embedding Anisotropy Diagnostic. ***Legend****: Pairwise cosine-similarity heatmaps for question 2 versus question 10 reasonings before (left) and after (right) PCA whitening. Even without whitening, unrelated questions separate to ~0.30-0.40 (≈65-70°), indicating the contrastively trained sentence model is already near-isotropic; whitening was therefore retained only as a sensitivity option and disabled in the primary analysis.*

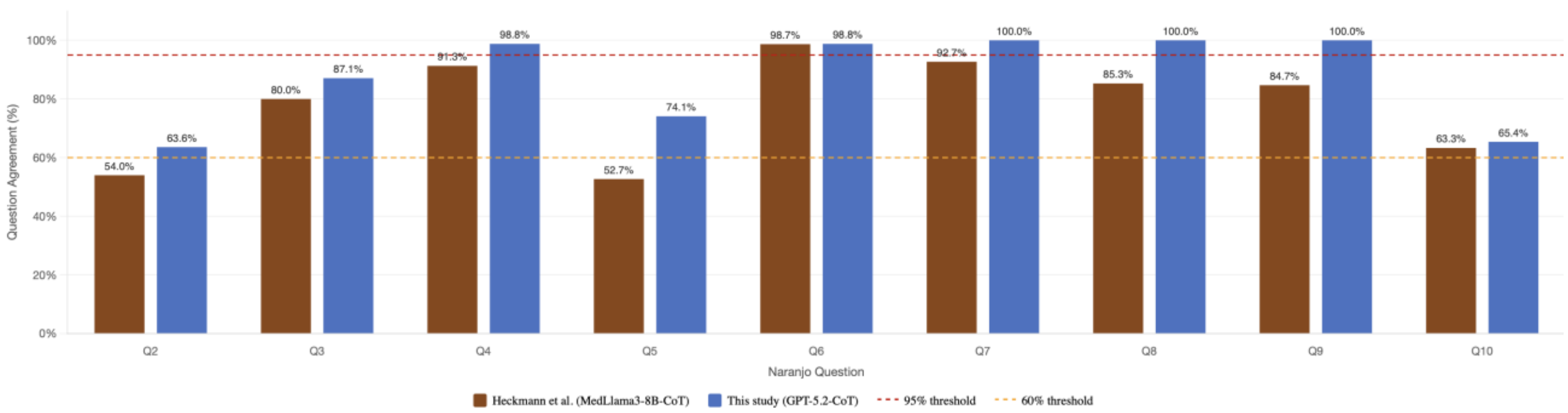

**Supplementary Figure S2.** Question and Final Classification Agreement: Comparison with Heckmann et al. ***Legend:*** *Question agreement rates for this study (GPT-5.2 with CoT, n = 723, blue) and Heckmann et al. (MedLlama3-8B with CoT, n = 150, brown). Dashed lines indicate acceptable agreement thresholds: 95% for questions 2-4 and 8 (red) and 60% for questions 5, 9, and 10 (orange). CoT = Chain-of-Thought; LLM = Large Language Model.*

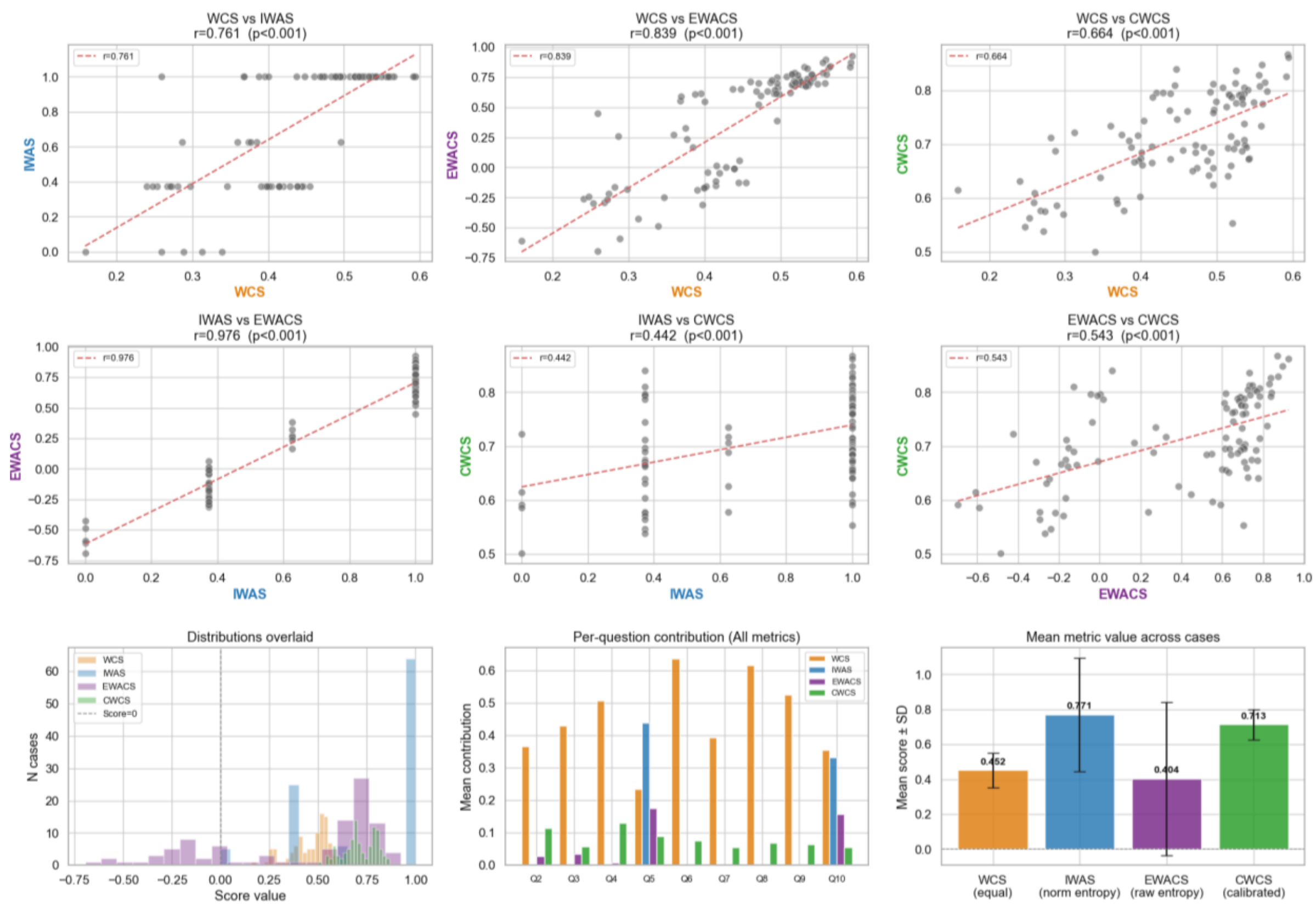

**Supplementary Figure S3.** Four-way comparison of WCS, IWAS, EWACS, and CWCS on the 100-case subset: pairwise scatter plots with Pearson correlations, overlaid distributions, per-question contributions, and mean ± SD.

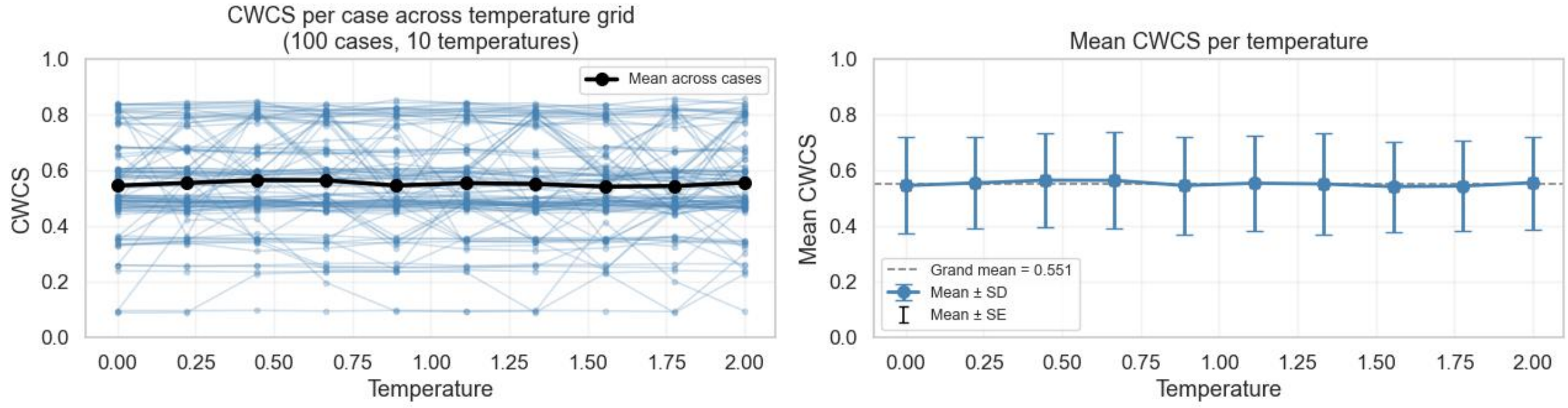

**Supplementary Figure S4.** Phase 1 CWCS temperature grid sweep (100 cases). ***Legend****: Mean CWCS ± SE across ten evenly spaced temperatures spanning [0, 2]; the response is essentially flat (range 0.023), with no systematic trend.*

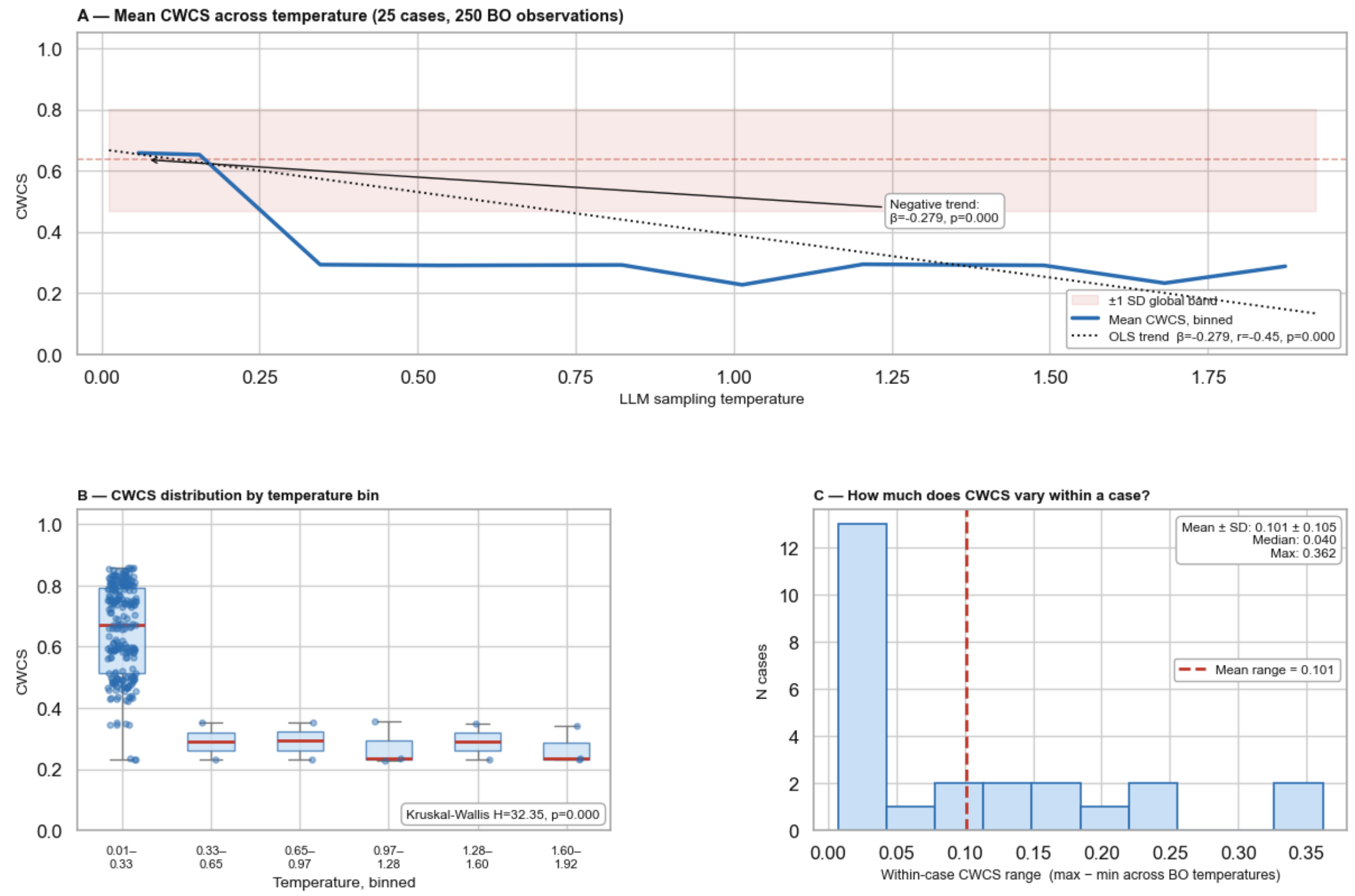

**Supplementary Figure S5.** Within-case (E1) temperature sensitivity under BO-guided selection (25 cases). ***Legend***: *Across repeated evaluations of the same case, CWCS shows a small but significant negative association with temperature ($\beta = -0.279$, $r = -0.45$, $p < 0.001$), i.e. lower temperatures tend to give marginally higher agreement for a fixed case.*

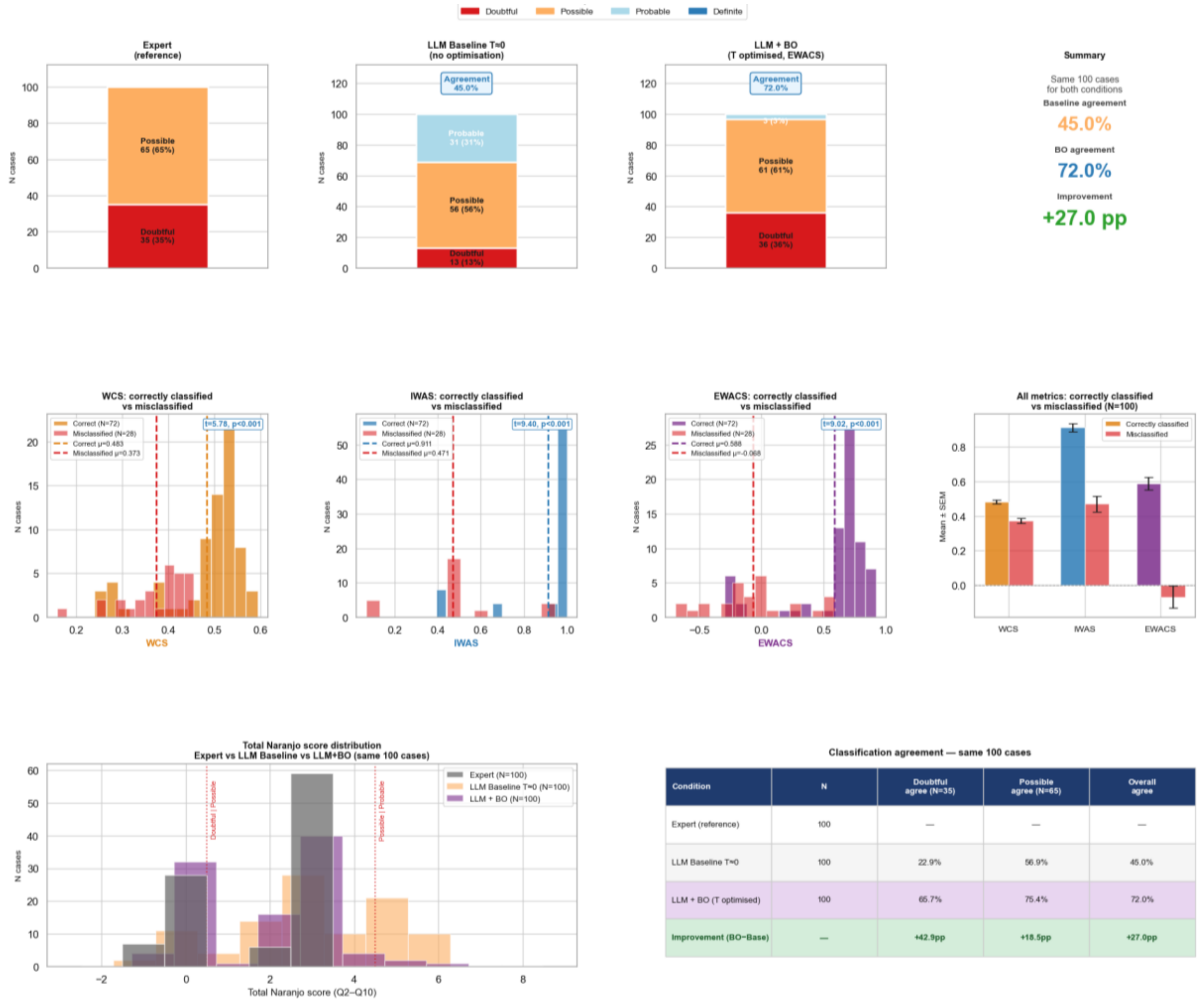

| Condition | N | Doubtful agree (N=35) | Possible agree (N=65) | Overall agree |
|---|---|---|---|---|
| Expert (reference) | 100 | — | — | — |
| LLM Baseline T≈0 | 100 | 22.9% | 56.9% | 45.0% |
| LLM + BO (T optimised) | 100 | 65.7% | 75.4% | 72.0% |
| Improvement (BO−Base) | — | +42.9pp | +18.5pp | +27.0pp |

**Supplementary Figure S6.** Naranjo classification: Expert vs. baseline LLM (T ≈ 0) vs. EWACS-optimized LLM on the same 100 cases. ***Legend***: *EWACS-guided optimization raises overall agreement from 45.0% to 72.0% (+27.0 pp), with the largest gains in Doubtful (+42.9 pp) and Possible (+18.5 pp) cases.*

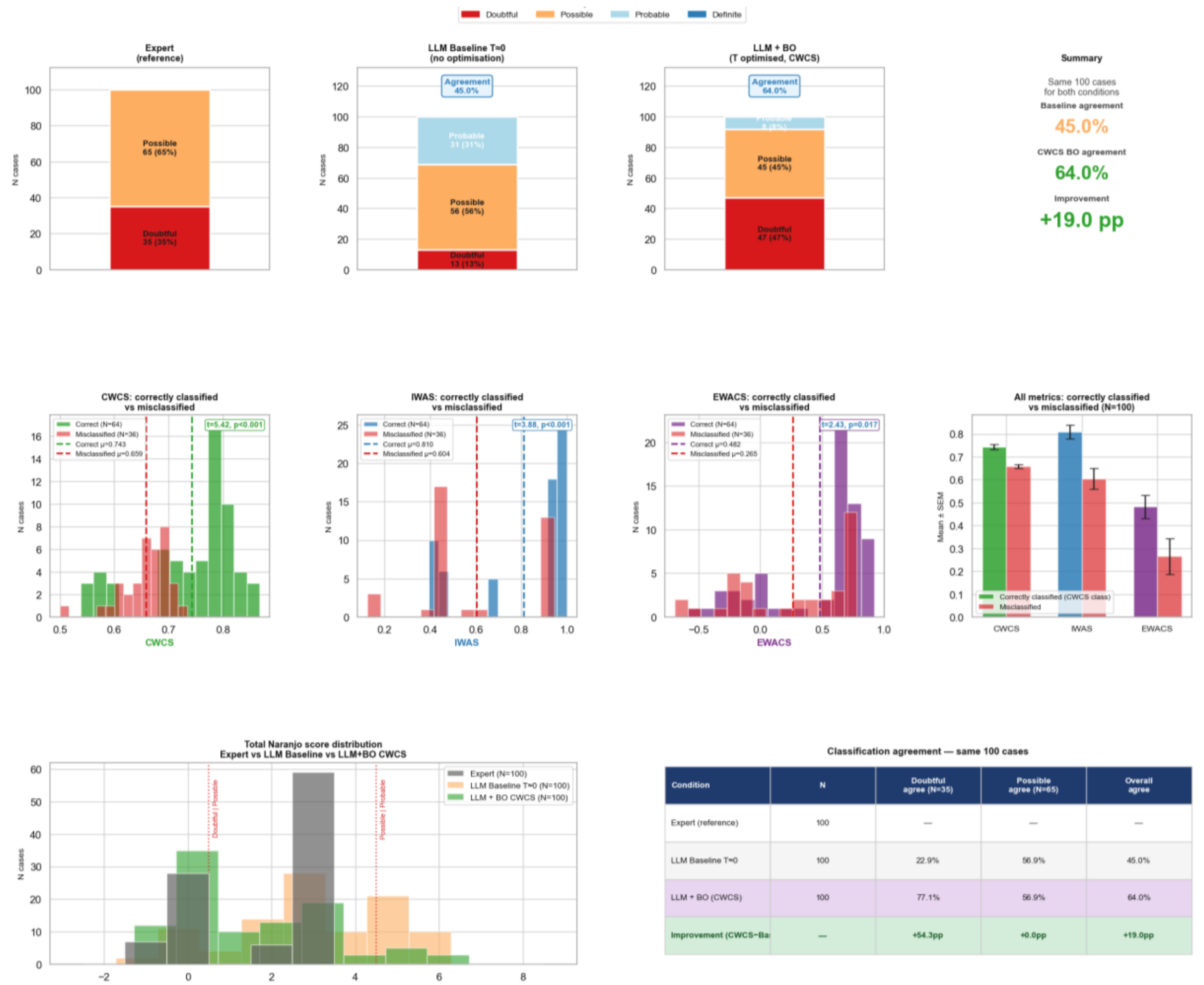

| Condition | N | Doubtful agree (N=35) | Possible agree (N=65) | Overall agree |
|---|---|---|---|---|
| Expert (reference) | 100 | — | — | — |
| LLM Baseline T=0 | 100 | 22.9% | 56.9% | 45.0% |
| LLM + BO (CWCS) | 100 | 77.1% | 56.9% | 64.0% |
| Improvement (CWCS−Ba | — | +54.3pp | +0.0pp | +19.0pp |

**Supplementary Figure S7.** Naranjo classification with the CWCS objective in place of EWACS (same 100 cases). ***Legend****: CWCS-guided optimization reaches 64.0% overall agreement, lower than EWACS, consistent with the finding that CWCS's reasoning-similarity component is noisier for the final classification target.*

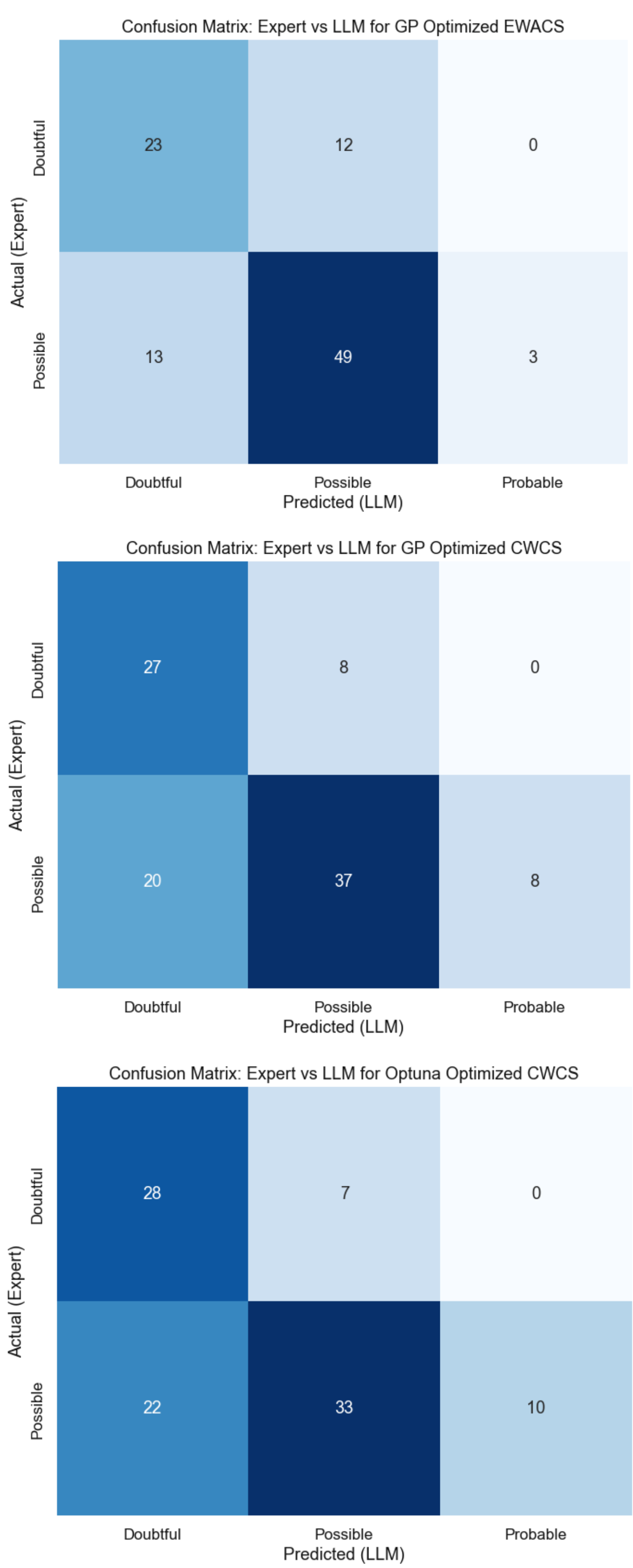

**Supplementary Figure S8.** Confusion matrices (Expert vs. optimized LLM) for the three optimization configurations on the same 100 cases: EWACS via GP (top; overall 72.0%), CWCS via GP (middle; 64.0%), and CWCS via TPE (bottom; 61.0%). Overall agreement equals the diagonal sum divided by 100.

## 7. Appendices

**Appendix 1.** Standard Operating Procedure for ICSR Assessments Using the Naranjo Algorithm

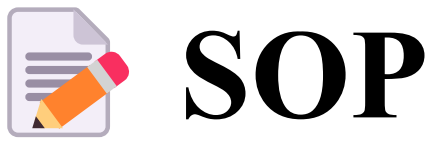

# SOP

| | |
|---|---|
| **Title** | Documentation of ICSR Assessments Using the Naranjo Algorithm |
| **Version** | 4.0 |
| **Date:** | February 6$^{th}$, 2026 |
| **Prepared by:** | Nicole Heckmann (kpr234@ku.dk), |
| **Reviewed by:** | Maurizio Sessa (maurizio.sessa@sund.ku.dk) |

# Table of Abbreviations

| | |
|---|---|
| **SOP** | Standard Operating Procedure |
| **ICSR** | Individual Case Safety Report |
| **FDA** | U.S. Food and Drug Administration |
| **FAERS** | FDA's Adverse Event Reporting System |
| **CSV** | Comma-Separated Values |
| **DEC** | Drug-Event Combination |
| **MedDRA** | Medical Dictionary for Regulatory Activities |
| **NDA** | New Drug Application |
| **PK** | Pharmacokinetic |
| **SQL** | Structured Query Language |

## 1. Purpose

This SOP describes the standardized procedure for documenting causality assessments of Individual Case Safety Reports (ICSRs) using the Naranjo Algorithm. The purpose is to ensure that all assessments are performed consistently and correctly, and that results are recorded in a uniform manner across cases.

## 2. Scope

This SOP applies to all Individual Case Safety Reports (ICSRs) that are assessed using the FDA's Adverse Event Case Viewer application.

# 3. The Naranjo Algorithm

The framework of the Naranjo Algorithm is described in *Table 1* below. [1]

**Table 1.** Naranjo Adverse Drug Reaction Probability Scale [1]

| ***Question*** | ***Yes*** | ***No*** | ***Do not know*** |
|---|---|---|---|
| 1. Are there previous conclusive reports on this reaction? | +1 | 0 | 0 |
| 2. Did the adverse event appear after the suspected drug was administered? | +2 | -1 | 0 |
| 3. Did the adverse event improve when the drug was discontinued, or was a specific antagonist administered? | +1 | 0 | 0 |
| 4. Did the adverse event reappear when the drug was readministered? | +2 | -1 | 0 |
| 5. Are there alternative causes that could on their own have caused the reaction? | -1 | +2 | 0 |
| 6. Did the reaction reappear when a placebo was given? | -1 | +1 | 0 |
| 7. Was the drug detected in blood or other fluids in concentrations known to be toxic? | +1 | 0 | 0 |
| 8. Was the reaction more severe when the dose was increased or less severe when the dose was decreased? | +1 | 0 | 0 |
| 9. Did the patient have a similar reaction to the same or similar drugs in any previous exposure? | +1 | 0 | 0 |
| 10. Was the adverse event confirmed by any objective evidence? | +1 | 0 | 0 |
| ***Total score:** ≥9 Definite[a]; 5-8 Probable[b]; 1-4 Possible[c]; ≤0 Doubtful[d]* | ***Total score:*** | | |

[a] A reaction was defined 'definite' if it was "one that (1) followed a reasonable temporal sequence after a drug or in which a toxic drug level had been established in body fluids or tissues, (2) followed a recognized response to the suspected drug, and (3) was confirmed by improvement on withdrawing the drug and reappeared on reexposure"

[b] A reaction was defined 'probable' if it was "(1) followed a reasonable temporal sequence after a drug, (2) followed a recognized response to the suspected drug, (3) was confirmed by withdrawal but not by exposure to the drug, and (4) could not be reasonably explained by the known characteristics of the patient's clinical state"

[c] A reaction was defined 'possible' if it was "(1) followed a temporal sequence after a drug, (2) possibly followed a recognized pattern to the suspected drug, and (3) could be explained by characteristics of the patient's disease"

[d] A reaction was defined 'doubtful' if it was "likely related to factors other than a drug"

For each of the ten Naranjo Algorithm questions (Q1-Q10), the expert:

- Assigns a numeric score according to the Naranjo guidelines [1].
- Provides reasoning, citing evidence from the case and/or external documents (details below).

After all questions are addressed, the expert calculates the Final Score by summing Q1-Q10 scores. The expert classifies the case into one of four categories as listed in *Table 2* below.

**Table 2.** Score Definitions [1]

| | | |
|---|---|---|
| $\geq 9$ | Definite Adverse Reaction | A reaction is defined 'definite' if it was "one that (1) followed a reasonable temporal sequence after a drug or in which a toxic drug level had been established in body fluids or tissues, (2) followed a recognized response to the suspected drug, and (3) was confirmed by improvement on withdrawing the drug and reappeared on reexposure" |
| 5-8 | Probable Adverse Reaction | A reaction is defined 'probable' if it was "(1) followed a reasonable temporal sequence after a drug, (2) followed a recognized response to the suspected drug, (3) was confirmed by withdrawal but not by exposure to the drug, and (4) could not be reasonably explained by the known characteristics of the patient's clinical state" |
| 1-4 | Possible Adverse Reaction | A reaction is defined 'possible' if it was "(1) followed a temporal sequence after a drug, (2) possibly followed a recognized pattern to the suspected drug, and (3) could be explained by characteristics of the patient's disease" |
| 0 or below | Doubtful Adverse Reaction | A reaction is defined 'doubtful' if it was "likely related to factors other than a drug" |

# 4. System Overview: FDA's Adverse Event Case Viewer

Link to app: https://caseviewer3.ngrok.app

### 4.1 Application Launch and Case Import

When the application is opened, the assessor is presented with the Data Source panel on the left-hand side.

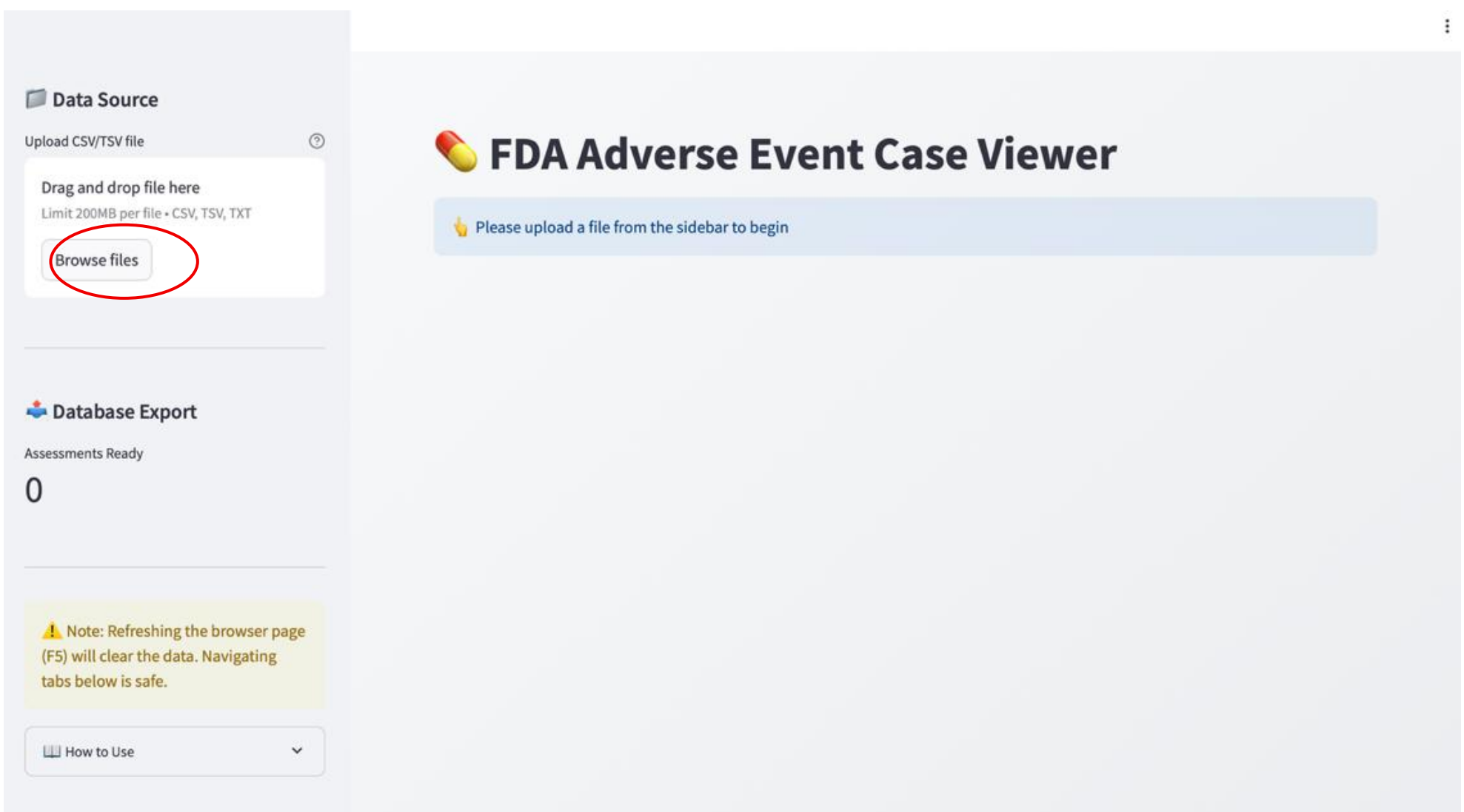

To begin work, the assessor must upload their allocated cases in CSV format by clicking the "Browse files" button in the left panel and selecting the assigned file.

- Once the file is successfully uploaded, the application confirms the import by displaying a message such as "✅ Loaded 100 cases" in the side panel.

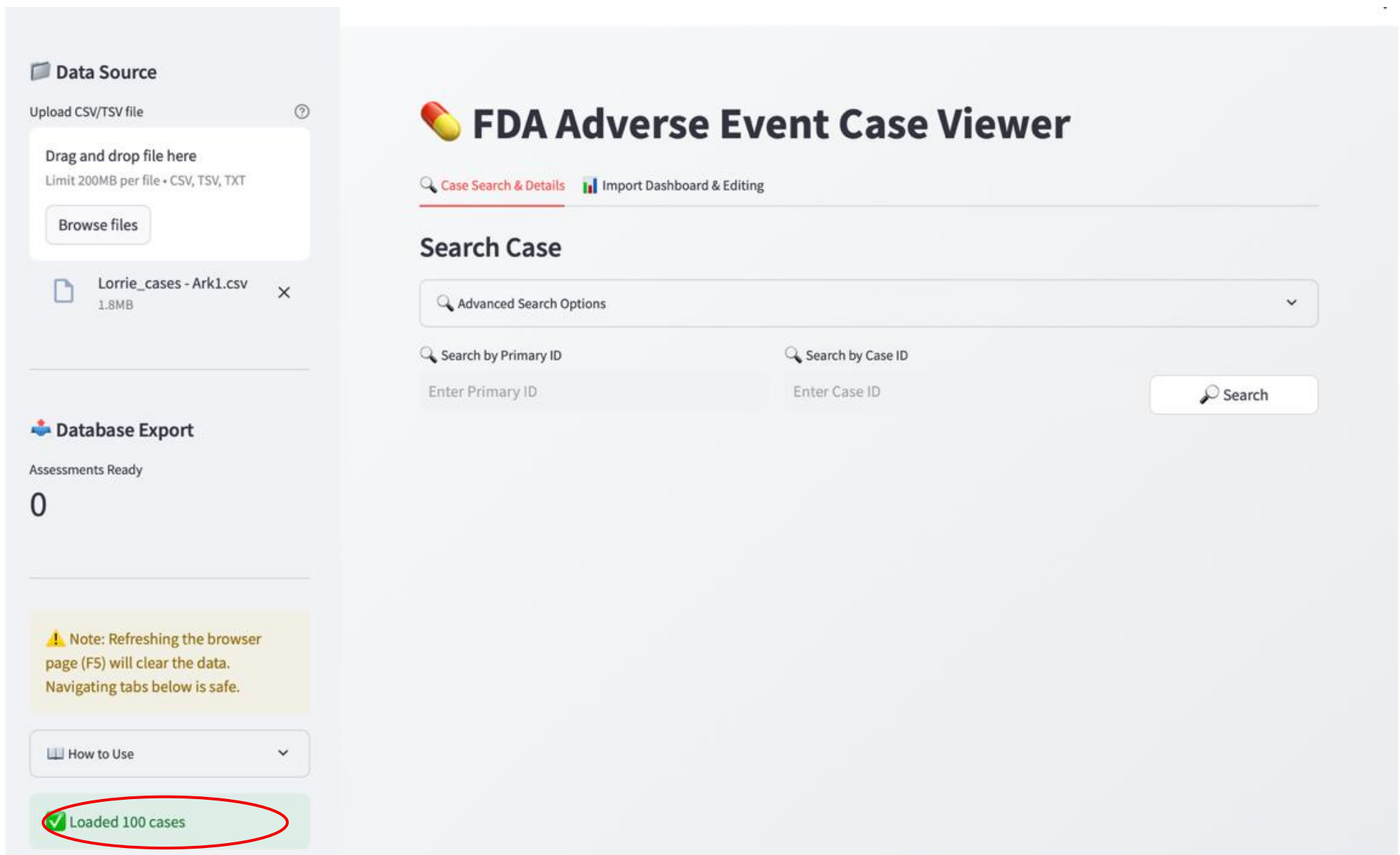

***Important operational note:*** Refreshing the browser page (F5) will clear all loaded data. Navigating between tabs within the application does not affect loaded data and is safe.

**4.2 Main Navigation Tabs**

The main interface consists of two primary tabs:

- 🔎 Case Search & Details → Used to retrieve and assess individual cases.
- 📊 Import Dashboard & Editing → Used to track overall progress and manage case status.

Assessors will move between these two tabs throughout the assessment process.

**4.3 Import Dashboard & Case Tracking**

Under the 📊 Import Dashboard & Editing tab, the assessor can view an overview of imported cases, including:

- Total imported cases
- Completed cases
- Assigned cases

Below this overview, the "Case List & Status Editor" table displays all cases with the following information:

- Case ID
- Primary ID
- Status
- Assessor
- Occurrence country

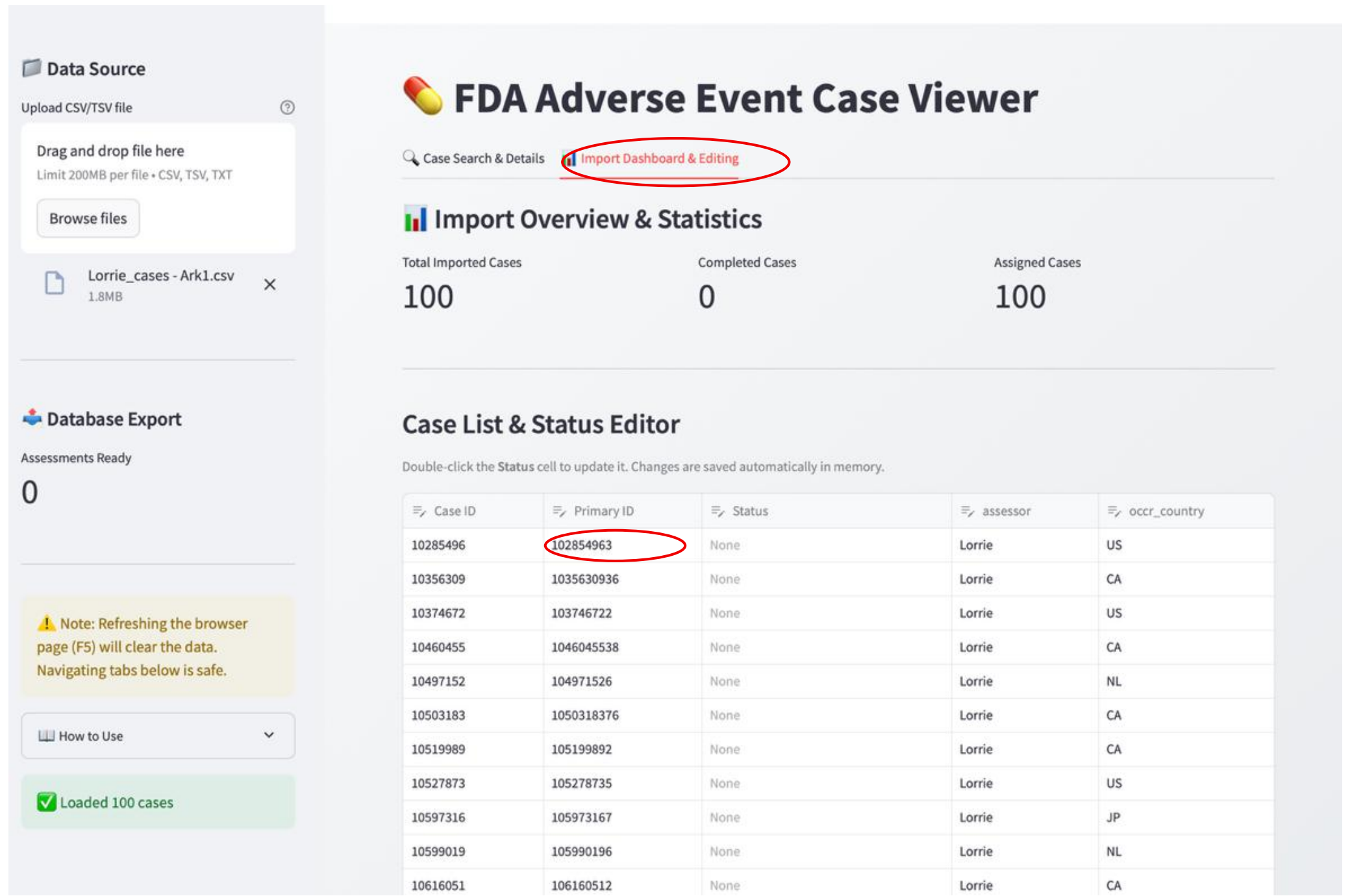

From this table, the assessor copies the Primary ID of the case to be assessed.

### 4.4 Case Retrieval

To retrieve a case for assessment:

1. Navigate to the 🔎 Case Search & Details tab.
2. Paste the copied Primary ID into the search field.
3. 🔎 Search

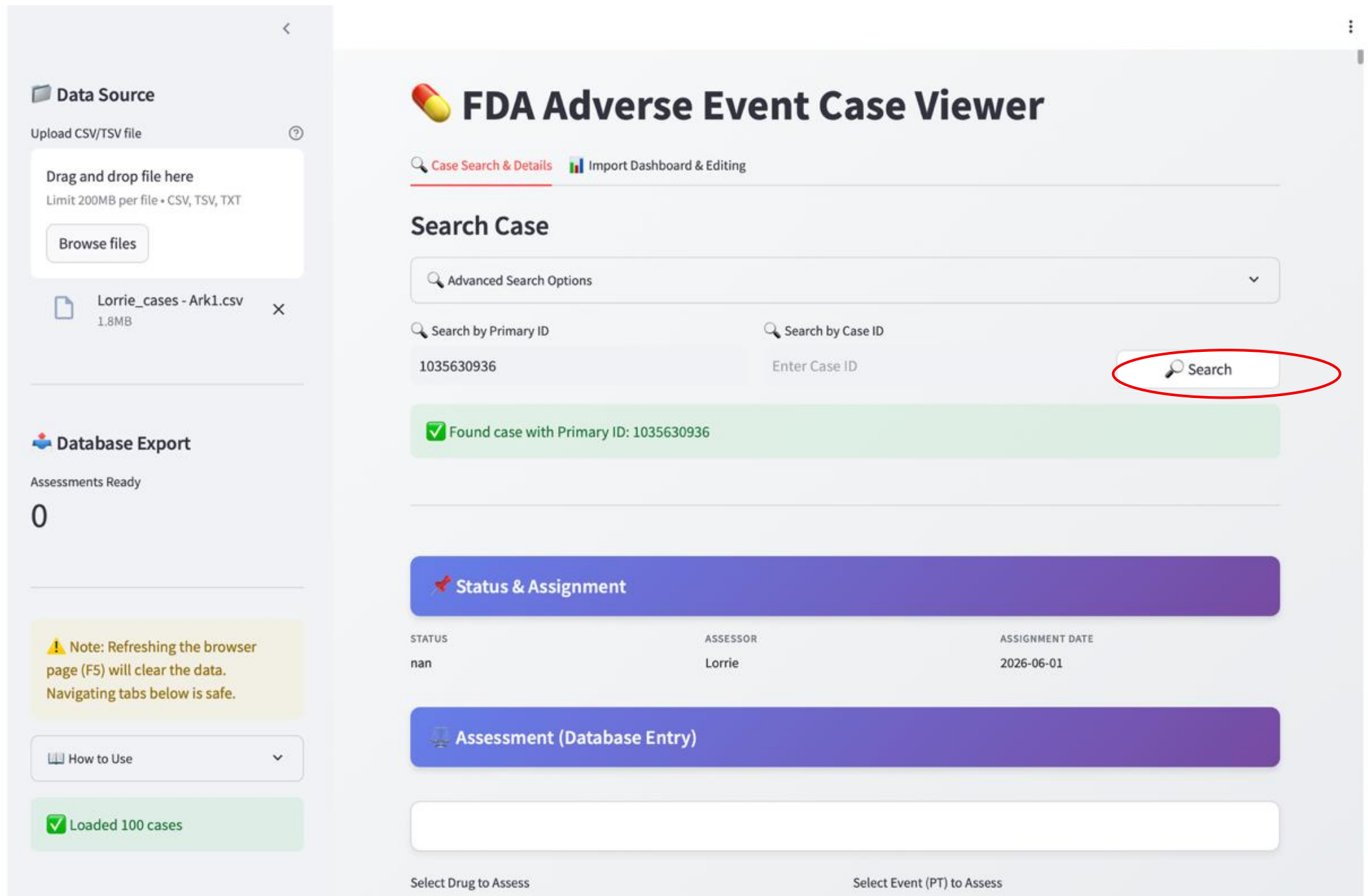

Once the case is successfully retrieved, the application displays a confirmation message: "✅ Found case with Primary ID: xxx", and the full case information becomes available on the page.

### 4.5 Status and Assignment Information

After a case is retrieved, the 📌 Status & Assignment section displays:

- Current case status
- Assigned assessor
- Assignment date

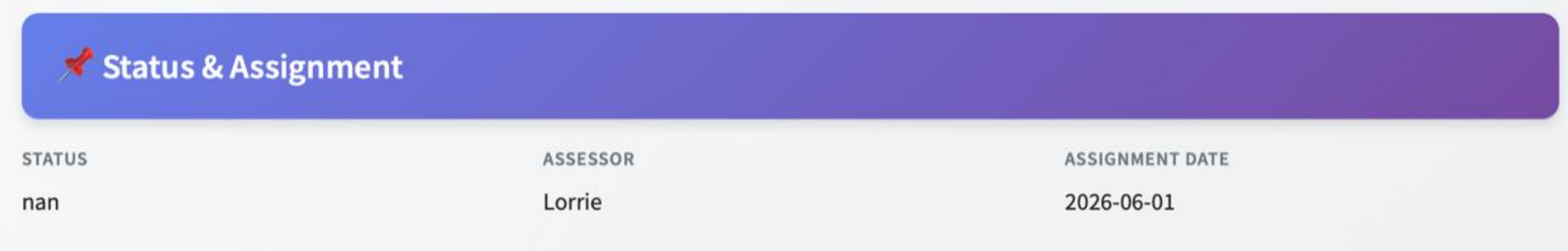

The status shown here updates automatically once the assessor marks the case as completed in the 📊 Import Dashboard & Editing tab.

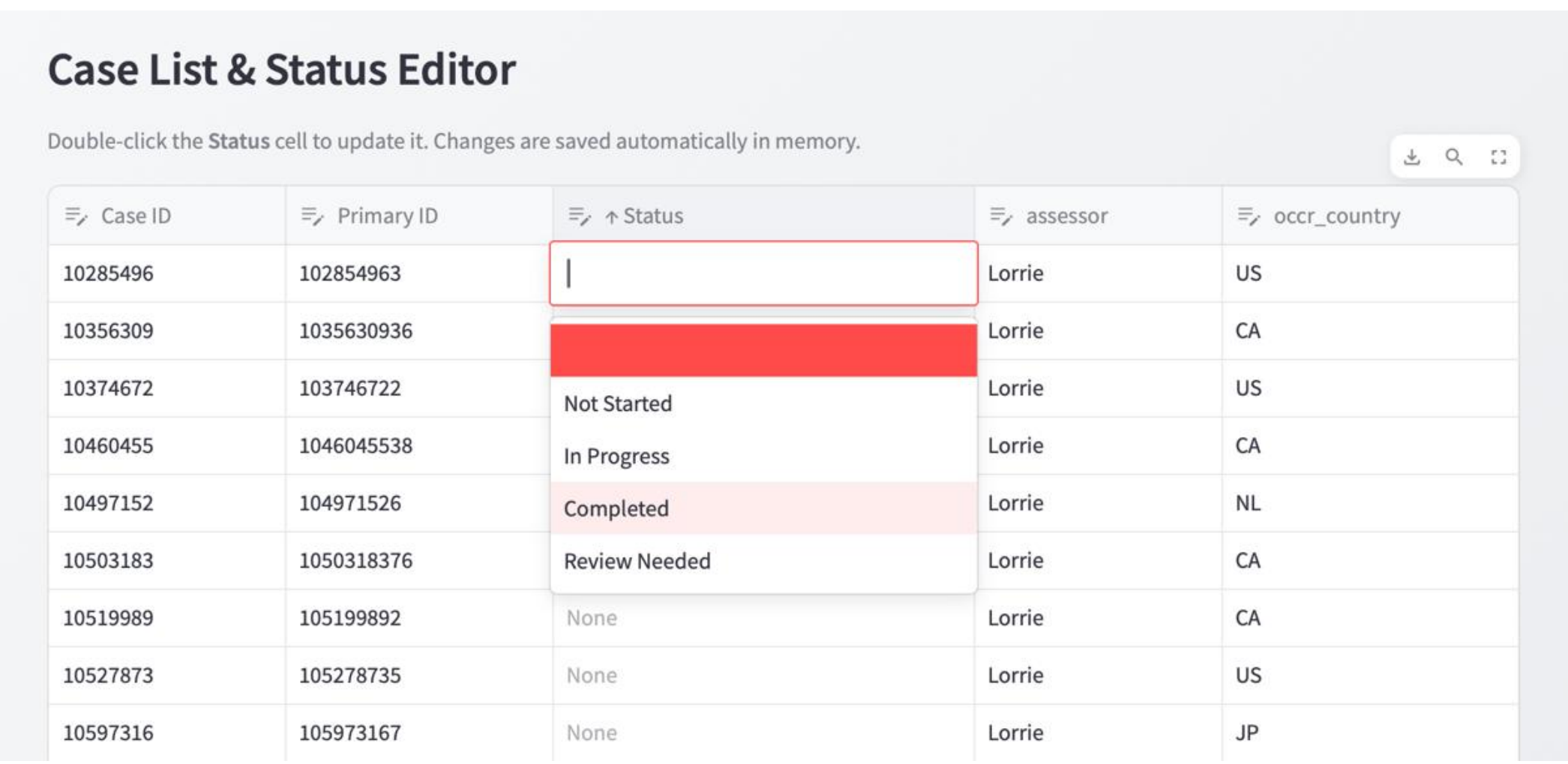
**Case List & Status Editor**

Double-click the **Status** cell to update it. Changes are saved automatically in memory.

| Case ID | Primary ID | ↑ Status | assessor | occr_country |
|---|---|---|---|---|
| 10285496 | 102854963 | | Lorrie | US |
| 10356309 | 1035630936 | | Lorrie | CA |
| 10374672 | 103746722 | Not Started | Lorrie | US |
| 10460455 | 1046045538 | In Progress | Lorrie | CA |
| 10497152 | 104971526 | Completed | Lorrie | NL |
| 10503183 | 1050318376 | Review Needed | Lorrie | CA |
| 10519989 | 105199892 | None | Lorrie | CA |
| 10527873 | 105278735 | None | Lorrie | US |
| 10597316 | 105973167 | None | Lorrie | JP |

### 4.6 Assessment Interface (Database Entry)

All causality assessments are performed in the Assessment (Database Entry) section.

Assessments are conducted one drug-event combination (DEC) at a time. For each assessment, the assessor must:

- Select one drug
- Select one adverse reaction (MedDRA Preferred Term)

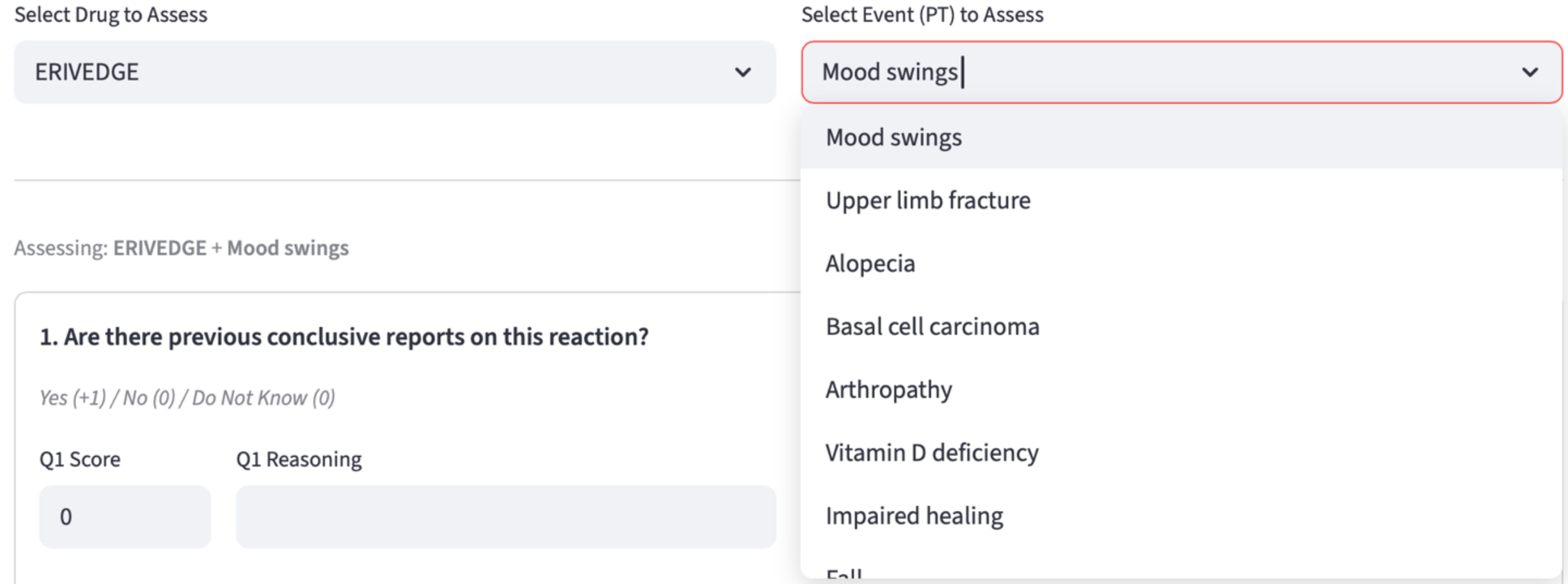

The assessor then answers Naranjo Algorithm Questions 1-10 directly in the interface, following the scoring rules and reasoning framework described in *Section 3* of this SOP (the process of answering the questions is described in *Section 5*).

Once all questions for the selected DEC are completed, the assessor clicks "💾 Save Assessment to Export Queue".

**1. Are there previous conclusive reports on this reaction?**

*Yes (+1) / No (0) / Do Not Know (0)*

| Q1 Score | Q1 Reasoning |
|---|---|
| 0 | not found in the FDA-label (link: https://dailymec |

**2. Did the adverse event appear after the suspected drug was administered?**

*Yes (+2) / No (-1) / Do Not Know (0)*

| Q2 Score | Q2 Reasoning |
|---|---|
| 0 | although event date and therapy start date are li |

**3. Did the adverse event improve when the drug was discontinued or a specific antagonist was administered?**

*Yes (+1) / No (0) / Do Not Know (0)*

| Q3 Score | Q3 Reasoning |
|---|---|
| 0 | the case does not provide any information on de |

**4. Did the adverse event reappear when the drug was readministered?**

*Yes (+2) / No (-1) / Do Not Know (0)*

| Q4 Score | Q4 Reasoning |
|---|---|
| 0 | the case does not provide any information on rec |

**5. Are there alternative causes that could on their own have caused the reaction?**

*Yes (-1) / No (+2) / Do Not Know (0)*

| Q5 Score | Q5 Reasoning |
|---|---|
| -1 | cancer can cause significant mood swings, rangi |

**6. Did the reaction reappear when a placebo was given?**

*Yes (-1) / No (+1) / Do Not Know (0)*

| Q6 Score | Q6 Reasoning |
|---|---|
| 0 | the case does not provide any information on pla |

**7. Was the drug detected in blood or other fluids in concentrations known to be toxic?**

*Yes (+1) / No (0) / Do Not Know (0)*

| Q7 Score | Q7 Reasoning |
|---|---|
| 0 | the case does not provide any information on pla |

**8. Was the reaction more severe when the dose was increased or less severe when the dose was decreased?**

*Yes (+1) / No (0) / Do Not Know (0)*

| Q8 Score | Q8 Reasoning |
|---|---|
| 0 | the case does not provide any PK data, thus the s |

**9. Did the patient have a similar reaction to the same or similar drugs in any previous exposure?**

*Yes (+1) / No (0) / Do Not Know (0)*

| Q9 Score | Q9 Reasoning |
|---|---|
| 0 | the case does not provide any information on ad |

**10. Was the adverse event confirmed by any objective evidence?**

*Yes (+1) / No (0) / Do Not Know (0)*

| Q10 Score | Q10 Reasoning |
|---|---|
| 0 | since the AE is a symptom (as opposed to a clinic |

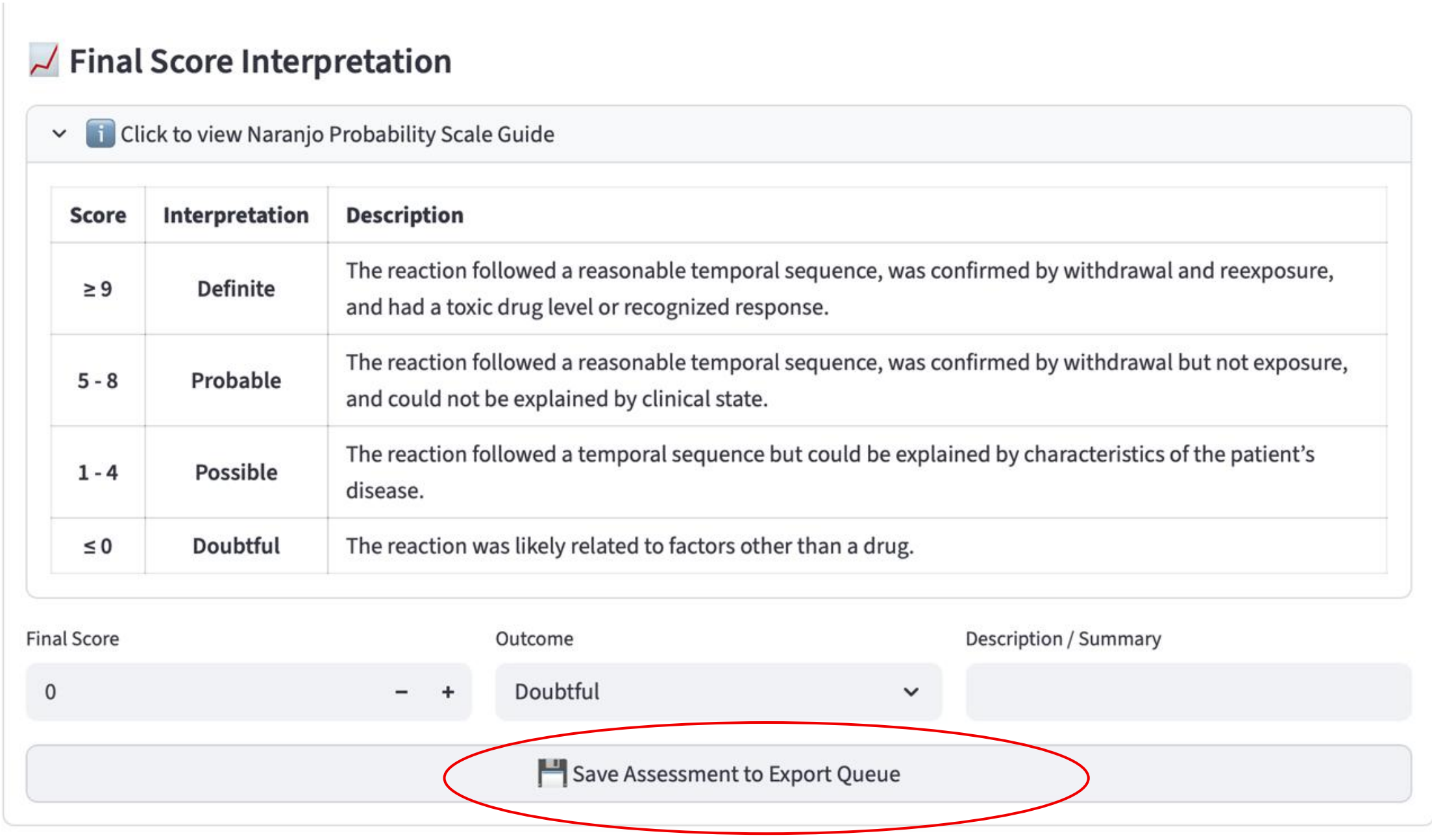

This process is repeated for each DEC within the case.

### 4.7 Supporting Case Information

Further down on the same page, the assessor has access to all supporting information extracted from FAERS, structured into the following sections:

- Administrative Information
- Patient Demographics
- Adverse Reactions
- Drug Information
- Case Narrative as extracted from FAERS (this is just the raw data, no need to use this)

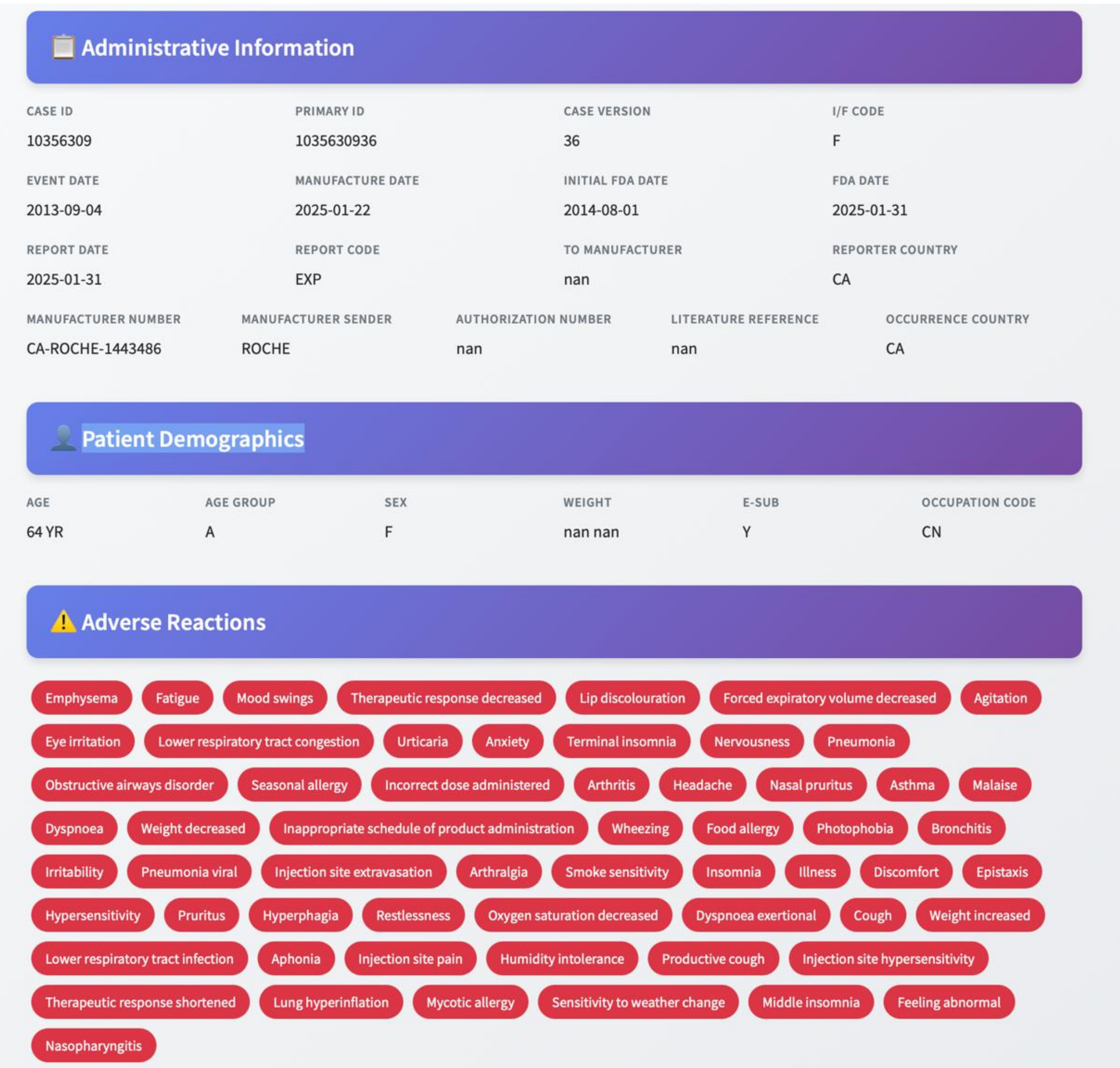

The Drug Information section is aligned per individual drug to avoid confusion.

- Under each listed drug, only information relevant to that specific drug is displayed. Duplicate drug entries have been removed so that only unique drugs are shown.

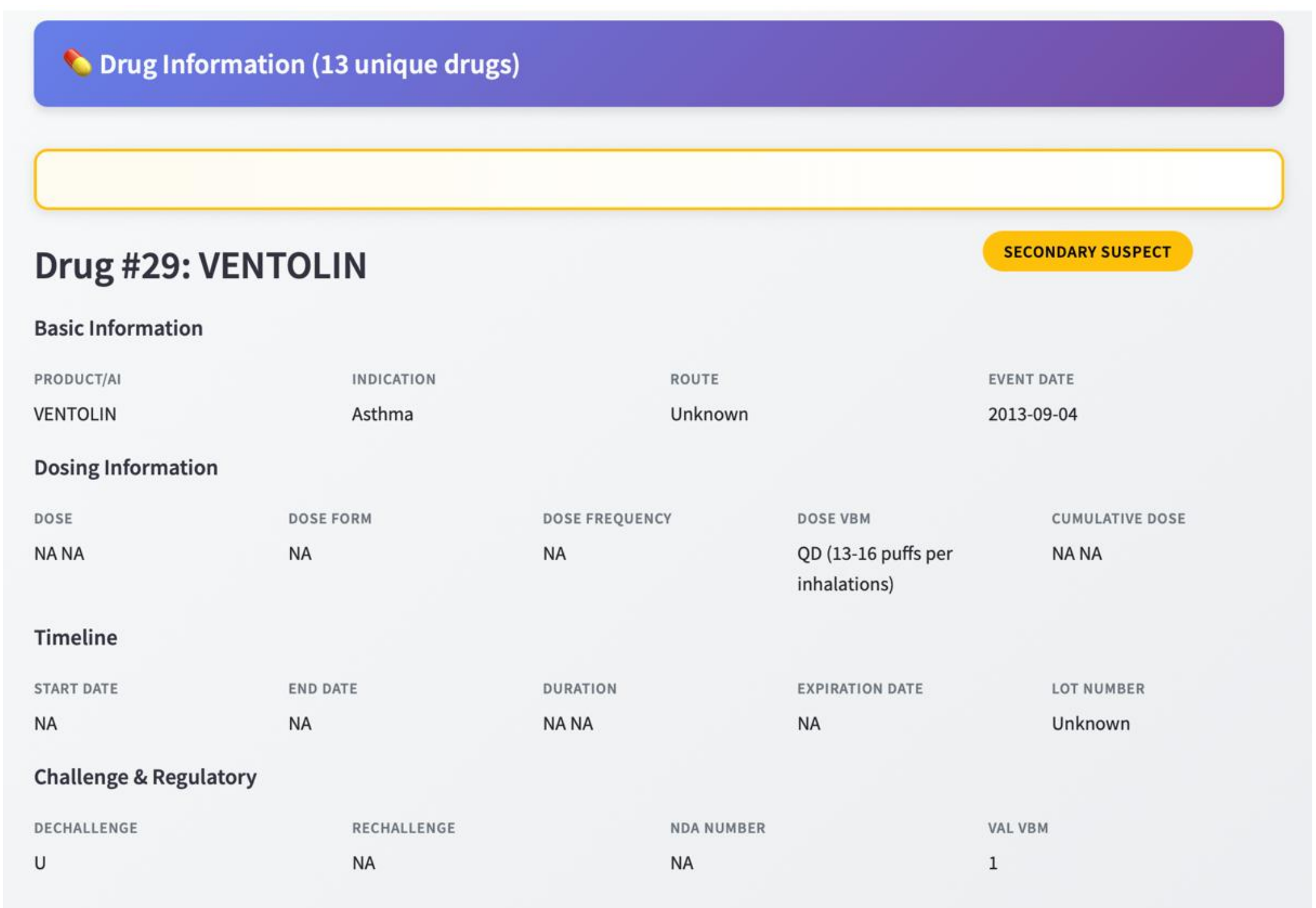
💊 Drug Information (13 unique drugs)

## Drug #29: VENTOLIN

SECONDARY SUSPECT

### Basic Information

| PRODUCT/AI | INDICATION | ROUTE | EVENT DATE |
| --- | --- | --- | --- |
| VENTOLIN | Asthma | Unknown | 2013-09-04 |

### Dosing Information

| DOSE | DOSE FORM | DOSE FREQUENCY | DOSE VBM | CUMULATIVE DOSE |
| --- | --- | --- | --- | --- |
| NA NA | NA | NA | QD (13-16 puffs per inhalations) | NA NA |

### Timeline

| START DATE | END DATE | DURATION | EXPIRATION DATE | LOT NUMBER |
| --- | --- | --- | --- | --- |
| NA | NA | NA NA | NA | Unknown |

### Challenge & Regulatory

| DECHALLENGE | RECHALLENGE | NDA NUMBER | VAL VBM |
| --- | --- | --- | --- |
| U | NA | NA | 1 |

## 5. Assessment & Documentation

Each ICSR is assessed using the ten Naranjo Algorithm questions (Q1-Q10). For each question, the expert assessor must:

a) assign a numeric score according to the Naranjo Algorithm [1]

b) provide written reasoning, citing evidence from the case and/or external references.

All scoring and reasoning are entered directly into the application's assessment window.

The scientific reasoning, interpretation rules, and decision logic for each question remain unchanged and are described below.

**Q1: Are there previous conclusive reports on this reaction?**

**Relevant ICSR information:** Adverse Reaction**,** Active Ingredient**,** Indication for Use.

**Process:**

1. Retrieve the latest FDA-label (Section 6. Adverse Reactions) from the FDA website (https://www.accessdata.fda.gov/scripts/cder/daf/index.cfm) by searching the active ingredient or brand name, opening the entry "Labels for NDA [NDA NUMBER]"; provide a direct hyperlink.
2. Search for the exact adverse reactions reported in the ICSR by reviewing Section 6, screening all listed adverse reactions, and checking for a perfect string match.
3. If not found, review synonyms, spelling variants, or lower-level terms using MedDRA as reference.
4. For FAERS adverse reactions related to exposure during pregnancy, Section 8. Use in Specific Populations (especially Section 8.1. Pregnancy) should be reviewed as well.
5. For FAERS adverse reactions related overdose, Section 10. Overdose should be reviewed as well.

**Example:** Adverse reaction reported as "stomach upset", but FDA-label lists "dyspepsia". Without MedDRA confirmation of synonymy, score 0.

**Perfect answer example:**

- Score: +1
- Reasoning: "Found in the FDA-approved label as nausea (link: https://www.accessdata.fda.gov/drugsatfda_docs/label/2021/020031s077lbl.pdf)"

**Q2: Did the adverse event appear after the suspected drug was administered?**

**Relevant ICSR information:** Event Date**,** Dose Frequency, Therapy Start/End Dates, Duration/Duration Code.

**Process:**

1. Review dosing timeline and event onset in the ICSR.
2. Compare administration date/time to reaction onset.
3. Confirm temporal plausibility (adverse reaction occurred after exposure, not before).
4. Confirm biological plausibility, i.e., the event type and latency must be compatible with known pharmacology or pathophysiology of the drug (e.g., allergic reactions typically require prior sensitization, hepatotoxicity usually develops after days to weeks).
5. If the event precedes drug intake, or the timing is biologically implausible, note this explicitly.

**Example:** Allergic rash reported after the first-ever dose, but true sensitization requires prior exposure; this may be coincidental.

**Perfect answer example:**

- Score: +2
- Reasoning: "Therapy start: 2022-11-28. Event onset: 2022-11-30. Rash developed two days post-exposure. → Score +2."

**Q3: Did the adverse reaction improve when the drug was discontinued (dechallenge)?**

**Relevant ICSR information:** Dechallenge.

**Process:**

1. Check for withdrawal of the drug in ICSR information.
2. Document whether reaction improved or resolved.

**Example:** The dechallenge field is coded as Y (*Yes*). This must be recorded as "The adverse event improved after dechallenge" and scored +1.

**Perfect answer example:**

- Score: +1
- Reasoning: "Dechallenge = Y. Rash improved after discontinuation. No other intervention. → Score +1."

**Q4: Did the adverse reaction reappear upon re-administration (rechallenge)?**

**Relevant ICSR information:** Rechallenge.

**Process:**

1. Check if drug was reintroduced.
2. Document recurrence of same reaction.

**Example:** Rechallenge coded as "NA" → must assign 0, not assume negative.

**Perfect answer example:**
- Score: +2
- Reasoning: "Rechallenge = NA → Score 0."

**Q5: Are there alternative causes (other than the drug) that could have caused the reaction?**

**Relevant ICSR information:** Patient Demographics/Medical History, Adverse Reaction, Active Ingredient, Posology, Indication.

**Process:**
1. Review concomitant medications recorded in the ICSR (Drug Sequence), noting any with known associations to the reported adverse reaction.
2. Review medical history and comorbidities (Patient demographics; narrative history if provided) for conditions that could plausibly explain the adverse reaction.
3. Evaluate the indication for use and underlying illness itself as a possible cause (e.g., infection causing rash, chemotherapy causing nausea).
4. Review posology information (Dose Amount, Dose Unit, Dose Form, Frequency, Route if available). Check if the adverse reaction could result from overdose, incorrect frequency, or inappropriate route, which may serve as an alternative explanation.
5. Consider reporter details (Occupation Code, Reporter Country) - sometimes relevant to context but not usually causal.

**Example:** Patient treated with azithromycin for infection; rash could plausibly be infection-related, not ibuprofen. Score −1.

**Perfect answer example:**
- Score: +2
- Reasoning: "No plausible alternative causes identified. No infection, no concomitant therapy except ibuprofen. → Score +2."

**Q6: Did the reaction reappear when a placebo was given?**

**Relevant ICSR information:** Typically, none in routine ICSRs.

**Example:** No placebo data mentioned. Must record "No data".

**Perfect answer example:**

- Score: 0
- Reasoning: "No placebo data provided. → Score 0."

**Q7: Was the drug detected in blood (or other fluids) at toxic concentrations?**

**Relevant ICSR information:** Typically, none. If provided, drug concentrations may appear in the narrative section.

**Example:** No PK data mentioned. Must record "No data".

**Perfect answer example:**

- Score: +1
- Reasoning: "No PK data provided. → Score 0."

**Q8: Was the reaction more severe when the dose was increased, or less severe when the dose was decreased?**

**Relevant ICSR information:** Adverse Reaction, Posology.

**Process:**

1. Examine dosing history.
2. Document correlation with severity.

**Tricky example:** Adverse reaction severity unchanged despite dose increase, must score 0.

**Perfect answer example:**

- Score: +1
- Reasoning: "Dose increased to 600 mg Q8H. Rash worsened accordingly. → Score +1."

**Q9: Did the patient have a similar reaction to the same or similar drugs in any previous exposure?**

**Relevant ICSR information:** Adverse Reaction, Active Ingredient, Indication. May appear in the narrative section.

**Process:**

1. Review patient history and prior exposures.
2. Note recurrence of same or similar event.

**Perfect answer example:**

- Score: +1

- Reasoning: “Patient reported rash after ibuprofen exposure in 2018 (documented in case history). → Score +1.”

**Q10: Was the adverse event confirmed by objective evidence?**

**Relevant ICSR information:** Adverse Reaction.

**Process:**

1. Review the adverse reaction.
2. Determine whether the adverse reaction is a sign or a symptom.
   - *Symptom = patient-reported (e.g., nausea, pain, itching, fatigue).*
   - *Sign = observed/verified objectively (e.g., rash noted by physician, abnormal lab value, hypertension)* [2]
3. If only a symptom is reported with no further indication, the answer is no/unknown, regardless of the source. [2]

**Example:** Patient reports “headache” without diagnostic confirmation. This is a symptom only → Score 0.

**Perfect answer example:**

- Score: +1
- Reasoning: “Rash erythematous observed and documented by physician. Objective sign confirmed by exam. → Score +1.”

**Final Documentation**

- **Final Score**: Sum of Q1-Q10 scores.
- **Outcome classification** (se detailed descriptions in *Table 2* [1]):

| | |
|---|---|
| ≥ 9 | Definite adverse reaction |
| 5-8 | Probable adverse reaction |
| 1-4 | Possible adverse reaction |
| 0 or below | Doubtful adverse reaction |

- A concise summary where the assessment outcome is integrated with the ICSR narrative. This should highlight the key clinical facts, relevant timelines, and the reasoning behind the final classification, ensuring the case can be understood without consulting the raw ICSR.

# 6. Export and Data Transfer

After completing all assessments, click "📥 Download Excel (Database Import)" found on the sidebar. The application generates a structured Excel file, with date and time stamp, containing all completed assessments.

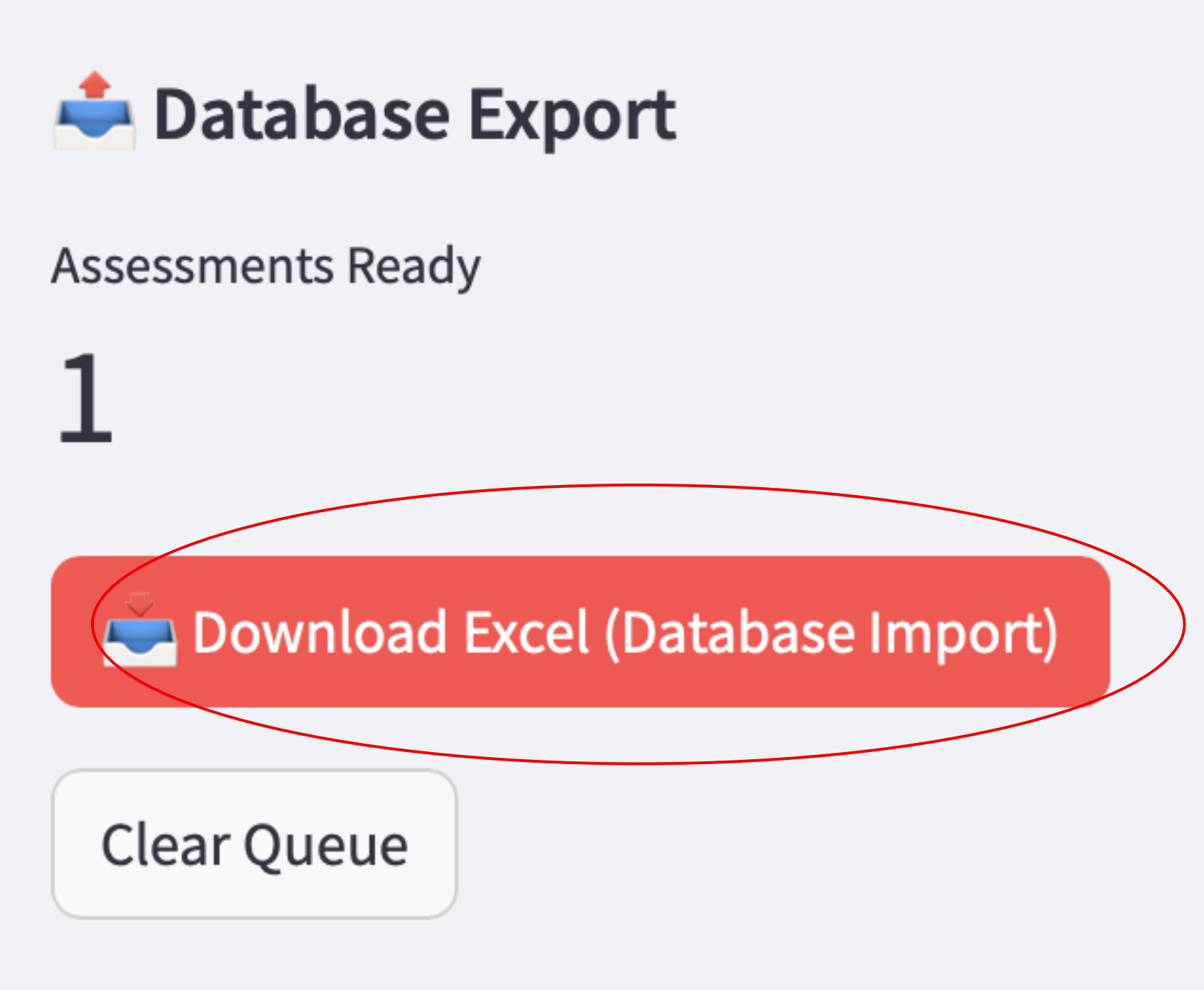

The assessor must send this file to:

Nicole Heckmann

📧 kpr234@ku.dk

Nicole will upload the file into the central SQL database.

**Appendix 2.** TRIPOD-LLM Checklist

| Section | Item | Description | Page no. |
|---|---|---|---|
| **Title** | 1 | Identify the study as developing, fine-tuning and/or evaluating the performance of an LLM, specifying the task, the target population and the outcome to be predicted. | Title page |
| **Introduction** | | | |
| **Abstract** | 2 | See TRIPOD-LLM for abstracts. | Abstract |
| **Background** | 3a | Explain the healthcare context/use case (for example, administrative, diagnostic, therapeutic and clinical workflow) and rationale for developing or evaluating the LLM, including references to existing approaches and models. | Section 1 |
| | 3b | Describe the target population and the intended use of the LLM in the context of the care pathway, including its intended users in current gold standard practices (for example, healthcare professionals, patients, public or administrators). | Section 1; Section 4.4; Section 4.5 |
| **Objectives** | 4 | Specify the study objectives, including whether the study describes the initial development, fine-tuning or validation of an LLM (or multiple stages). | Section 1; Abstract |
| **Methods** | | | |
| **Data** | 5a | Describe the sources of data separately for the training, tuning and/or evaluation datasets and the rationale for using these data (for example, web corpora, clinical research/trial data, EHR data or unknown). | Section 2.1 |
| | 5b | Describe the relevant data points and provide a quantitative and qualitative description of their distribution and other relevant descriptors of the dataset (for example, source, languages and countries of origin). | Sections 2.1, 3.1; Figure 1 |
| | 5c | Specifically state the date of the oldest and newest item of text used in the development process (training, fine-tuning and reward modeling) and the evaluation datasets. | Section 2.3; Section 2.9 |
| | 5d | Describe any data preprocessing and quality checking, including whether this was similar across text corpora, institutions and relevant sociodemographic groups. | Sections 2.1, 2.2; Appendix 1 |
| | 5e | Describe how missing and imbalanced data were handled and provide reasons for omitting any data. | Sections 2.1, 3.2, 4.1, 4.6 |
| **Analytical Methods** | 6a | Report the LLM name, version and last date of training. | Section 2.3 |
| | 6b | Report details of the LLM development process, such as LLM architecture, training, fine-tuning procedures and alignment strategy (for example, reinforcement learning and direct preference optimization) and alignment goals (for example, helpfulness, honesty and harmlessness). | Section 2.3 |
| | 6c | Report details of how the text was generated using the LLM, including any prompt engineering (including consistency of outputs), and inference settings (for example, seed, temperature, max token length and penalties), as relevant. | Sections 2.3, 2.6, 2.9; Supplementary Table S1 |
| | 6d | Specify the initial and postprocessed output of the LLM (for example, probabilities, classification and unstructured text). | Sections 2.3, 2.4 |
| | 6e | Provide details and rationale for any classification and, if applicable, how the probabilities were determined and thresholds identified. | Sections 2.4, 2.8; Supplementary Tables S2, S3 |
| **LLM Output** | 7a | Include metrics that capture the quality of generative outputs, such as consistency, relevance, accuracy and presence/type of errors compared to gold standards. | Sections 2.4, 2.5, 2.8; Table 1; Figure 4c |
| | 7b | Report the outcome metrics' relevance to the downstream task at deployment time and, where applicable, the correlation of metric to human evaluation of the text for the intended use. | Section 3.4; Figure 4c |
| | 7c | Clearly define the outcome, how the LLM predictions were calculated (for example, formula, code, object and API), the date of inference for closed-source LLMs and evaluation metrics. | Sections 2.3-2.5; Equations 2-6 |
| | 7d | If outcome assessment requires subjective interpretation, describe the qualifications of the assessors, any instructions | Section 2.2; Appendix 1 |
| | | provided, relevant information on demographics of the assessors and inter-assessor agreement. | |
| | 7e | Specify how performance was compared to other LLMs, humans and other benchmarks or standards. | Section 3.2; Section 4.3; Supplementary Figure S2 |
| **Annotation** | 8a | If annotation was done, report how the text was labeled, including providing specific annotation guidelines with examples. | Section 2.2; Appendix 1 |
| | 8b | If annotation was done, report how many annotators labeled the dataset(s), including the proportion of data in each dataset that was annotated by more than one annotator, and the inter-annotator agreement. | Section 2.2 |
| | 8c | If annotation was done, provide information on the background and experience of the annotators or the characteristics of any models involved in labeling. | Section 2.2 |
| **Prompting** | 9a | If research involved prompting LLMs, provide details on the processes used during prompt design, curation and selection. | Sections 2.3; Supplementary Table S1 |
| | 9b | If research involved prompting LLMs, report what data were used to develop the prompts. | Section 2.3; Supplementary Table S1 |
| **Summariza-tion** | 10 | Describe any preprocessing of the data before summarization. | N/A |
| **Instruction tuning/ alignment** | 11 | If instruction tuning/alignment strategies were used, what were the instructions, data and interface used for evaluation, and what were the characteristics of the populations doing the evaluation? | N/A |
| **Compute** | 12 | Report compute, or proxies thereof (for example, time on what and how many machines, cost on what and how many machines, inference time, floating-point operations per second), required to carry out methods. | Section 2.9 |
| **Ethical approval** | 13 | Name the institutional research board or ethics committee that approved the study and describe the participant-informed consent or the ethics committee waiver of informed consent. | N/A |
| **Open science** | 14a | Give the source of funding and the role of the funders for the present study. | N/A |
| | 14b | Declare any conflicts of interest and financial disclosures for all authors. | N/A |
| | 14c | Indicate where the study protocol can be accessed or state that a protocol was not prepared. | N/A |
| | 14d | Provide registration information for the study, including register name and registration number, or state that the study was not registered. | N/A |
| | 14e | Provide details of the availability of the study data. | N/A |
| | 14f | Provide details of the availability of the code to reproduce the study results. | N/A |
| **Public involvement** | 15 | Provide details of any patient and public involvement during the design, conduct, reporting, interpretation or dissemination of the study or state no involvement. | N/A |
| **Results** | | | |
| **Participants** | 16a | When using patient/EHR data, describe the flow of text/EHR/patient data through the study, including the number of documents/questions/participants with and without the outcome/label and follow-up time as applicable. | Section 3.1; Figure 1 |
| | 16b | When using patient/EHR data, report the characteristics overall and for each data source or setting and development/evaluation splits, including the key dates, key characteristics and sample size. | Section 3.1; Figure 1 |
| | 16c | For LLM evaluation that includes clinical outcomes, show a comparison of the distribution of important clinical variables that may be associated with the outcome between development and evaluation data, if available. | Section 3.1; Figure 1 |
| | 16d | When using patient/EHR data, specify the number of participants and outcome events in each analysis (for example, for LLM development, hyperparameter tuning and LLM evaluation). | Sections 3.2, 3.4 |
| **Performance** | 17 | Report LLM performance according to prespecified metrics (see item 7a) and/or human evaluation (see item 7d). | Sections 3.2-3.4; Table 1; Figures 2, 3, 4; Supplementary Figures S2-S8 |
| **LLM updating** | 18 | If applicable, report the results from any LLM updating, including the updated LLM and subsequent performance. | N/A |
| **Discussion** | | | |
| **Interpretation** | 19a | Give an overall interpretation of the main results, including issues of fairness in the context of the objectives and previous studies. | Sections 4.1-4.4 |
| **Limitations** | 19b | Discuss any limitations of the study and their effects on any biases, statistical uncertainty and generalizability. | Section 4.6 |
| **Usability of the LLM in context** | 19c | Describe any known challenges in using data for the specified task and domain context with reference to representation, missingness, harmonization and bias. | Sections 4.1, 4.6 |
| | 19d | Define the intended use for the implementation under evaluation, including the intended input, end-user and level of autonomy/human oversight. | Sections 4.4, 4.5 |
| | 19e | If applicable, describe how poor quality or unavailable input data should be assessed and handled when implementing the LLM; that is, what is the usability of the LLM in the context of current clinical care. | N/A |
| | 19f | If applicable, specify whether users will be required to interact in the handling of the input data or use of the LLM, and what level of expertise is required of users. | N/A |
| | 19g | Discuss any next steps for future research, with a specific view of the applicability and generalizability of the LLM. | Section 4.5 |

***Legend:*** *M = LLM methods; D = de novo LLM development; E = LLM evaluation; H = LLM evaluation in healthcare settings; C = classification; OF = outcome forecasting; QA = long-form question answering; IR = information retrieval; DG = document generation; SS = summarization and simplification; MT = machine translation; API = application programming interface.*

**Appendix 3.** CHART Checklist

| Heading | No | Chart checklist item | Page no. |
|---|---|---|---|
| **Title and Abstract** | | | |
| **Title** | 1a | State that the study is assessing one or more generative AI-driven chatbots for clinical evidence or health advice. | Title page |
| **Abstract/ Summary** | 1b | Apply a structured format, if applicable. | Abstract |
| **Introduction** | | | |
| **Background** | 2a | State the scientific background, rationale, and healthcare context for evaluating the generative AI-driven chatbot(s), referencing relevant literature when applicable. | Section 1 |
| | 2b | State the aims and research questions including the target audience, intervention, comparator(s), and outcome(s). | Section 1; Abstract |
| **Methods** | | | |
| **Model Identifiers** | 3a | State the name and version identifier(s) of the generative AI model(s) and chatbot(s) under evaluation, as well as their date of release or last update. | Section 2.3 |
| | 3b | State whether the generative AI model(s) and chatbot(s) are open-source or closed-source/proprietary. | Section 2.3 |
| **Model Details** | 4a | State whether the generative AI model was a base model or a novel base model, tuned model, or fine-tuned model. | Section 2.3 |
| | 4b | If a base model is used, cite its development in sufficient detail to identify the model. | Section 2.3; Section 2.9 |
| | 4c | If a novel base model, tuned model, or fine-tuned model is used, describe the pre- and/or post-implementation/deployment data and parameters. | N/A |
| **Prompt Engineering** | 5a | Describe the evolution of study prompt development. | Section 2.3 |
| | 5ai | Describe the sources of prompts. | Section 2.3; Supplementary Table S1 |
| | 5aii | State the number and characteristics of the individual(s) involved in prompt engineering. | N/A |
| | 5aiii | Provide details of any patient and public involvement during prompt engineering. | N/A |

| | 5b | Provide study prompts. | Supplementary Table S1 |
|---|---|---|---|
| **Query Strategy** | 6a | State route of access to generative AI model. | Section 2.3; Section 2.9 |
| | 6b | State the date(s) and location(s) of queries for the generative AI-driven chatbot(s) including the day, month, and year as well as city and country. | Section 2.3 |
| | 6c | Describe whether prompts were input into separate chat session(s). | Section 2.3 |
| | 6d | Provide all generative AI-driven chatbot output/responses | N/A |
| **Prompting** | 7a | Define the ground truth or reference standard used to define successful generative AI-driven chatbot performance. | Section 2.2; Appendix 1 |
| | 7b | Describe the process undertaken for generative AI-driven chatbot performance evaluation. | Sections 2.4-2.8 |
| | 7bi | State the number and characteristics of team members involved in performance evaluation. | Section 2.2; Appendix 1 |
| | 7bii | Provide details of any patients and public involvement during the evaluation process. | N/A |
| | 7biii | State whether evaluators were blinded to the identity of the generative AI-driven chatbot(s) under assessment. | N/A |
| **Sample Size** | 8 | Report how the sample size was determined. | Sections 2.1, 2.6 |
| **Data Analysis** | 9a | Describe statistical analysis methods, including any evaluation of reproducibility of generative AI-driven chatbot responses. | Section 2.8; Section 2.9 |
| | 9ai | Report the measures used for performance evaluation. | Table 1; Sections 3.2-3.4; Figures 2-4 |
| **Results** | | | |
| | 10a | Report the performance evaluation undertaken including the alignment between generative AI-driven chatbot output and ground truth or reference standard using quantitative or mixed methods approaches as applicable. | Sections 3.2-3.4; Table 1; Figures 2, 3, 4; Supplementary Figures S2, S6-S8 |
| | 10b | For responses deviating from the ground truth or reference standard, state the nature of the difference(s). | Section 3.2; Figure 2 |
| | 10c | Indicate where the study protocol can be accessed or state that a protocol was not prepared. | N/A |
| **Discussion** | | | |
| | 11a | Interpret study findings in the context of relevant evidence. | Sections 4.3, 4.4 |
| | 11b | Describe the strengths and limitations of the study. | Section 4.6 |
| | 11c | Describe the potential implications for practice, education, policy, regulation, and research. | Sections 4.4, 4.5 |
| **Open Science** | | | |
| **Disclosures** | 12a | Report any relevant conflicts of interest for all authors. | N/A |
| **Funding** | 12b | Report sources of funding and their role in the conduct and reporting of the study. | N/A |
| **Ethics** | 12c | Describe the process undertaken for ethical approval. | N/A |
| | 12ci | Describe the measures taken to safeguard data privacy of patient health information, as applicable. | N/A |
| | 12cii | State whether permission/licensing was obtained for the use of original, copyrighted data. | N/A |
| **Protocol** | 12d | Provide a study protocol. | N/A |
| **Data availability** | 12e | State where study data, code repository, and model parameters can be accessed. | N/A |